\documentclass[10pt,twocolumn,letterpaper,final]{article}
\usepackage{wacv}
\usepackage{times}
\usepackage{epsfig}
\usepackage{multirow}
\usepackage{graphicx}
\usepackage{amsmath}
\usepackage{amssymb}
\usepackage{array}
\usepackage{commath}
\usepackage{amsthm}
\usepackage{algorithmic}
\usepackage{breqn}

\usepackage[pagebackref=true,breaklinks=true,colorlinks,bookmarks=false]{hyperref}

\def\wacvPaperID{} 

\wacvfinalcopy 

\ifwacvfinal
\fi
\usepackage{fancyhdr}

\fancypagestyle{firstpage}{%
  \rhead{IEEE ICME 2022}
  \lfoot{\small{This paper is accepted in IEEE International Conference on Multimedia and Expo (ICME), 2022 and copyright has been transferred to IEEE.
  Personal use of this material is permitted. Permission from IEEE must be obtained for all other users, including reprinting/ republishing this material for advertising or promotional purposes, creating new collective works for resale or redistribution to servers or lists, or reuse of any copyrighted components of this work in other works.}}
}

\begin{document}

\title{Vision Transformer Hashing for Image Retrieval}

\author{Shiv Ram Dubey, Satish Kumar Singh\\
\textit{Computer Vision and Biometrics Lab} \\\textit{Indian Institute of Information Technology, Allahabad, India}\\
{\small srdubey@iiita.ac.in, sk.singh@iiita.ac.in}
\and
Wei-Ta Chu\\
\textit{Department of Computer Science and Information Engineering} 
\\\textit{National Cheng Kung University, Taiwan}\\
{\small wtchu@gs.ncku.edu.tw}
}

\maketitle
\thispagestyle{firstpage}

\begin{abstract}
Deep learning has shown a tremendous growth in hashing techniques for image retrieval. Recently, Transformer has emerged as a new architecture by utilizing self-attention without convolution. Transformer is also extended to Vision Transformer (ViT) for the visual recognition with a promising performance on ImageNet. In this paper, we propose a Vision Transformer based Hashing (VTS) for image retrieval. We utilize the pre-trained ViT on ImageNet as the backbone network and add the hashing head. The proposed VTS model is fine tuned for hashing under six different image retrieval frameworks, including Deep Supervised Hashing (DSH), HashNet, GreedyHash, Improved Deep Hashing Network (IDHN), Deep Polarized Network (DPN) and Central Similarity Quantization (CSQ) with their objective functions. We perform the extensive experiments on CIFAR10, ImageNet, NUS-Wide, and COCO datasets. The proposed VTS based image retrieval outperforms the recent state-of-the-art hashing techniques with a great margin. We also find the proposed VTS model as the backbone network is better than the existing networks, such as AlexNet and ResNet. The code is released at \url{https://github.com/shivram1987/VisionTransformerHashing}.
\end{abstract}

\section{Introduction}
Deep learning has witnessed a revolution for more than a decade for several applications of computer vision, natural language processing, and many more \cite{deeplearning}. Deep learning learns the problem-specific important features automatically from data in an hierarchical fashion with different levels of abstractness. Different types of Neural Networks have been utilized to deal with different types of data. Specifically, Convolutional Neural Networks (CNNs) are used for image data \cite{AlexNet}. Whereas, Recurrent Neural Networks (RNNs) are used for sequential data \cite{rnn}.

Several CNN models have been investigated for different computer vision applications, such as image classification \cite{AlexNet},
\cite{ResNet}, 
object detection \cite{fastrcnn}, \cite{fasterrcnn}, segmentation
\cite{maskrcnn}, \cite{panet}, face recognition \cite{taigman2014deepface}, 
and medical applications \cite{liu2021subtype}.
CNNs have also been very heavily exercised for image hashing and retrieval \cite{dubey2021decade}. Deep Supervised Hashing (DSH) \cite{dsh} is one of the early attempts to utilize CNN by quantizing the network outputs to binary hash code. The DSH uses a regularizer on the real-valued network outputs to produce the discrete binary values. Tanh function based continuation method is incorporated in HashNet \cite{hashnet} for smooth transition from real-valued features to binary codes. The HashNet also utilizes the weighted cross-entropy loss to learn the sparse data in a similarity-preserving manner. The gradients are passed as identity mapping for the hash layer to avoid the vanishing gradients in GreedyHash method \cite{greedyhash}. The GreedyHash uses the Sign function in the hash layer. Improved Deep Hashing Network (IDHN) \cite{idhn} utilizes the cross-entropy loss and mean square error loss to deal with `hard similarity’ and `soft similarity’, respectively, for multi-label image retrieval. Central Similarity Quantization (CSQ) \cite{csq} relies on the optimization of the central similarity between data points in terms of their hash centers to increase the distinctiveness of the hash codes for image retrieval. Deep Polarized Network (DPN) \cite{dpn} utilizes polarization loss as bit-wise hinge-like loss which leads to the different output channels to be far away from zero. Inherently, DPN brings a high level of separability between the hash codes of different classes. These models utilize CNNs as the backbone network.

\begin{figure*}[!t]
    \centering
    \includegraphics[trim={26 17 100 25},clip,width=\textwidth]{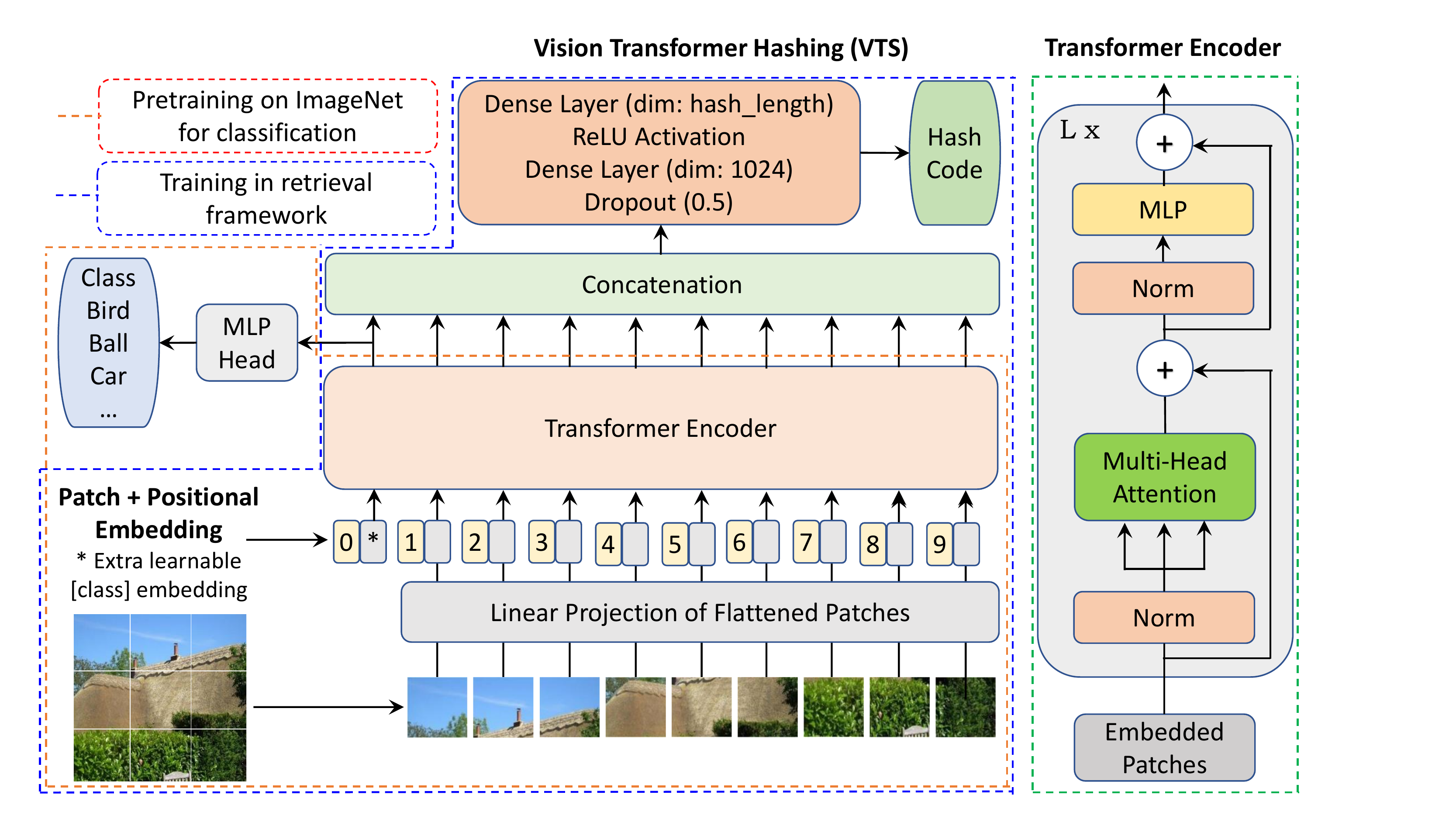}
    \caption{Vision Transformer Hashing (VTS) Framework for Image Retrieval in Blue Dotted Line using the Pre-trained Vision Transformer (ViT Model) \cite{vit} for Image Classification in Red Dotted Line.}
    \label{fig:vts}
\end{figure*}

Though the CNNs have been very successful, it requires a deep network to achieve the global representation of the image in feature space. However, at the other end, the RNNs are able to utilize the entire data in each step in the feature representation. But, the RNNs are not parallelizable as the input at a given time-step is dependent on the output of the previous time-step. Transformer networks are the recent trends in deep learning with very promising performance \cite{tay2020efficient}, \cite{lin2021survey}. The transformers tackle the downsides of both CNNs and RNNs by modeling the global representation of the input feature at each step with highly parallelizable architecture. Basically, the transformers utilize the self-attention modules \cite{transformer}. The self-attention mechanism captures the relative importance within the same input sequence in the feature representation of that sequence. 

The transformers have also witnessed its impact over computer vision problems \cite{khan2021transformers}, \cite{han2020survey}. The Vision Transformer (ViT) \cite{vit} is the state-of-the-art to utilize the transformer for image recognition at scale. The ViT considers the patches of $16 \times 16$ of the image as the sequential input to the transformers. The ViT has shown outstanding performance as compared to the CNNs for visual recognition. Very recently, the vision transformers have gained a huge popularity for several vision tasks such as visual tracking \cite{wang2021transformer}, \cite{chen2021transformer}, 
dense prediction \cite{ranftl2021vision}, 
medical image segmentation \cite{li2021medical}, person re-identification \cite{li2021diverse}, text-to-visual retrieval \cite{miech2021thinking}, and human-object interaction detection \cite{kim2021hotr}, among others.

A very few attempts have been made to use the transformer for image hashing \cite{li2021self}, \cite{gkelios2021investigating}, \cite{el2021training}, \cite{tan2021instance}, \cite{transhash}. 
Bidirectional transformer \cite{li2021self} utilizes the bidirectional correlations between frames for video hashing. However, the bidirectional transformer does not utilize the vision transformer. 
Transformer is used as an off-the-shelf feature extractor in \cite{gkelios2021investigating}.
Vision transformer is trained in \cite{el2021training} for image retrieval with a metric learning objective function utilizing the contrastive loss on real-valued descriptors. The hashing is not performed in \cite{el2021training}. 
Similarly, Reranking Transformer is used for image retrieval in \cite{tan2021instance} without hashing.
TransHash \cite{transhash} utilizes a siamese vision transformer for feature learning. However, the TransHash is not able to utilize the pre-training of Vision Transformers over large-scale dataset and the recently investigated improved objective functions for image hashing.

Motivated from the success of the transformers for different problems, we propose a Vision Transformer Hashing (VTS) for image retrieval in this paper. We utilize the pre-trained ViT \cite{vit} models to extract the real-valued features and fine-tune in several image hashing setup with additional hashing module. We use six state-of-the-art hashing frameworks with the proposed VTS model as the backbone network. An outstanding performance is observed using the proposed VTS model on several benchmark image retrieval datasets. The contributions of are as follows:
\begin{itemize}
    \item We propose an end-to-end trainable Vision Transformer Hashing (VTS) model for image retrieval by utilizing the pre-trained ViT models \cite{vit}.
    \item The proposed VTS model utilizes the ViT model as a generic feature extractor, removes the MLP head, appends a hashing module and uses the existing hashing frameworks to train the model.
    \item Extensive experiments have been conducted to show the excellent performance by the VTS model on benchmark CIFAR10, ImageNet, NUS-Wide and MS-COCO datasets under six state-of-the-art hashing frameworks. 
\end{itemize}

The remaining paper is structured as follows: Section 2 introduces the Vision Transformer Hashing (VTS); Section 3 presents the experimental setup; Section 4 discusses the results; and finally Section 5 sets the concluding remarks.

\section{Proposed Vision Transformer Hashing}
The proposed Vision Transformer Hashing (VTS) framework is illustrated in Fig. \ref{fig:vts}. It consists of patch embedding generation, transformer encoder, pre-training under classification framework and hashing module for image retrieval.

\subsection{Patch Embeddings Generation}
The input image $I \in \mathbb{R}^{m, m, c}$ is divided into $N$ non-overlapping patches $I_i \in \mathbb{R}^{k, k, c}$ where $i = 1,2,...,N$, $m^2 = Nk^2$ and $c = 3$ is the number of color channels. The input patches $I_i$ are flattened into vectors $V_i \in \mathbb{R}^{1,d}$ where $d = {ck^2}$ and $i = 1,2,...,N$. A linear projection is performed on each flattened vector to generate the patch embeddings ($PE$) as,
\begin{equation}
    PE_i = V_i \times W_{PE}
\end{equation}
where $V_i \in \mathbb{R}^{1,d}$ is the flattened vector corresponding to the $i^{th}$ patch, $W_{PE} \in \mathbb{R}^{d,de}$ is the parameter matrix and $PE_i \in \mathbb{R}^{1,de}$ is the projected embeddings corresponding to the $i^{th}$ patch. Here, $de$ is the embedding dimension (i.e., hidden size) of the projection. Thus, the projected embeddings corresponding to the entire input image can be written as $PE \in \mathbb{R}^{N,de}$.
A class token embedding $CT \in \mathbb{R}^{1,de}$ is introduced as the learnable parameters initialized with zeros and concatenated with the projected embeddings in order to generate the expanded embeddings ($EE \in \mathbb{R}^{N+1,de}$) as $EE = [CT, PE]$ on $1^{st}$ dimension.
In order to incorporate the spatial information, the position embedding ($PoE \in \mathbb{R}^{N+1,de}$) is included as the learnable parameters initialized with zeros and added in the expanded embeddings. So, the final projection embedding ($FE$) is given as,
\begin{equation}
    FE = dropout(EE + PoE, 0.1)
\end{equation}
where $dropout$ is a dropout layer with $0.1$ dropout factor and $FE \in \mathbb{R}^{N+1,de}$ is the final projection embeddings to be given as the input to transformer encoder.




\subsection{Transformer Encoder}
The transformer encoder consists of the stack of $L$ transformer blocks. 
A transformer block transforms the given hidden-state input into an output of the same dimension with the help of a self-attention mechanism. 
The input to a transformer block $T_j|_{j = 1,2,...,L}$ is the hidden state $h_{j-1}$ which is the output of previous transformer block $T_{j-1}$, where $h_{j-1} \in  \mathbb{R}^{N+1,de}$. The input to the first transformer block $T_{1}$ is the final projection embedding (i.e., $h_0 = FE$). The output of the last transformer block $T_L$ is hidden state $h_L$. Basically, the input and output of the $j^{th}$ transformer block are denoted by $h_{j-1}$ and $h_j$, respectively. 

The layer normalization is the first layer in the $j^{th}$ transformer block which takes input $h_{j-1}$ and produces output $l_{j-1} \in \mathbb{R}^{N+1,de}$. The output of layer normalization is the input to self-attention layer which consists of three parametric projections for Query, Key and Value vectors using the learnable weight matrices $W_Q$, $W_K$ and $W_V$, respectively. The output of linear projections are
\begin{equation}
    S_t = l_{j-1} \times W_t
\end{equation}
where $S_t \in \mathbb{R}^{N+1,de}$ for $t = \{Q,K,V\}$. As the attention layer uses multi-attention heads, the Query ($S_Q$) and Value ($S_V$) vectors are reshaped to a dimension of $(A_h,N+1,A_S)$ and the Key vactor is reshaped to the dimension of $(A_h,A_S,N+1)$ where $A_h$ is the number of attention heads and $A_S$ is the attention head size which is computed as $A_S = de/A_h$. Next, a matrix multiplication is performed between $S_Q$ and $S_K$ to produce an output matrix $S_{QK} \in \mathbb{R}^{A_h,N+1,N+1}$ which is normalized by square root of attention head size as,
\begin{equation}
    S_{QK} = \frac{S_Q \times S_K}{\sqrt{A_S}}.
\end{equation}
Next, the attention weights ($A_w$) are computed using softmax function on $S_{QK}$ as,
\begin{equation}
    A_w = softmax(S_{QK}).
\end{equation}
The Value vector ($S_V$) is operated with the attention weights ($A_w$) to generate the attention weighted features ($F_A \in \mathbb{R}^{A_h,N+1,A_S}$) as,
\begin{equation}
    F_A = A_w \times S_V
\end{equation}
which is further reshaped in the dimension of $(N+1,de)$, i.e., $F_A \in \mathbb{R}^{N+1,de}$.
Finally, a linear projection is applied on $F_A$ with a learnable parameter $W_A$ to produce the output of the self-attention module ($F_S$) as $F_S = F_A \times W_A$. 
In order to improve the training, the transformer blocks include the residual connection as,
\begin{equation}
    F_{R} = h_{j-1} + F_S
\end{equation}
where $h_{j-1}$ is the input to transformer block, $F_S$ is the output of self-attention module and $F_R$ is the output of residual connection.
A layer normalization is then used on the output of residual connection $F_{R}$ with the output of this layer as $l_{R}$ of same dimensionality.
A multilayer perceptron module is applied on the output of layer normalization with a linear projection to $mlp_{dim} = 3072$ dimension, GELU activation function, dropout layer with $0.1$ dropout factor and a linear projection to $de = 768$ dimension. 
Finally, a residual connection is used to produce the output of the transformer block as,
\begin{equation}
    h_j = F_R + F_{MLP}
\end{equation}
where $F_{MLP} \in \mathbb{R}^{N+1,de}$ is the output of the MLP module, $F_R$ is the output of the first residual connection and $h_j$ is the output of the $j^{th}$ transformer block. Similarly, the output of the last transformer block $h_L$ is also computed. The final output of the transformer encoder (i.e., $TE \in \mathbb{R}^{N+1,de}$) is the layer normalization output on $h_L$ which is passed to the next step, such as pre-training under classification framework and hashing under retrieval framework (see Fig. \ref{fig:vts}).

\subsection{Vision Transformer Pretraining}
In this paper, we use the pretrained Vision Transformer (ViT) model \cite{vit}. The pretraining is performed using the final output of the transformer encoder on the ImageNet dataset under classification framework. The ViT extracts the representative features $RF \in \mathbb{R}^{1,de}$ from the first vector of transformer encoder output $TS$. The ViT uses a multilayer perceptron (MLP) head which computes the class scores from the $RF$ features. We use two versions of pretrained ViT, including ViT-B\_32 and ViT-B\_16 which use the patch sizes of 32 and 16, respectively. We use the code and pretrained models available from GitHub open repository\footnote{https://github.com/jeonsworld/ViT-pytorch} for hashing.

\subsection{Hashing Module}
We train the pre-trained vision transformer model under image retrieval framework. In order to facilitate the image retrieval and learning of hash code, we add a hash block on top of the transformer encoder output. First, the output of transformer encoder $TE$ is reshaped to a dimension of $((N+1) \times de)$ which is passed through a dropout layer with $0.5$ dropout factor. Then a linear projection is applied to transform the feature into a dimension of $1024$ followed by the ReLU activation function. Finally, a linear projection is applied to generate the final hash features ($H_f$) having the number of values the same as the hash bit length. Note that we use six state-of-the-art methods to generate the hash code from hash features. Based on the input patch size, i.e., $32$ and $16$, we propose two VTS variants, i.e., VTS32 and VTS16, respectively.

In this paper, we test the VTS performance under six state-of-the-art hashing frameworks, namely Deep Supervised Hashing (DSH) \cite{dsh}, HashNet \cite{hashnet}, GreedyHash \cite{greedyhash}, Improved Deep Hashing Network (IDHN) \cite{idhn}, Central Similarity Quantization (CSQ) \cite{csq} and Deep Polarized Network (DPN) \cite{dpn}. We use the hash features $H_f$ as the input to these frameworks and train the entire network including transformer encoder and hash module using the objective function of the corresponding hashing framework. Basically, the vision transformer with hash module (i.e., VTS) can be seen as the backbone network under these hashing frameworks. The code for these hashing frameworks is considered from the open GitHub repository\footnote{https://github.com/swuxyj/DeepHash-pytorch}. The experiments are also carried out using AlexNet \cite{AlexNet} and ResNet50 \cite{ResNet} backbone networks to show the improved performance using the proposed VTS32 and VTS16 backbone networks.

\begin{table*}[!t]
\caption{The retrieval results in terms of mAP in \% for 16, 32 and 64 bit hash codes using AlexNet, ResNet and VTS backbone networks under DSH \cite{dsh}, HashNet \cite{hashnet}, GreedyHash \cite{greedyhash}, IDHN \cite{idhn}, CSQ \cite{csq} and DPN \cite{dpn} frameworks on CIFAR-10, ImageNet, NUS-Wide and MS-COCO datasets. The best results are highlighted in bold.}
    \centering
    \begin{tabular}{m{0.077\textwidth}|m{0.026\textwidth}m{0.026\textwidth}m{0.026\textwidth}|m{0.026\textwidth}m{0.026\textwidth}m{0.026\textwidth}|m{0.026\textwidth}m{0.026\textwidth}m{0.026\textwidth}|m{0.026\textwidth}m{0.026\textwidth}m{0.026\textwidth}|m{0.026\textwidth}m{0.026\textwidth}m{0.026\textwidth}|m{0.026\textwidth}m{0.026\textwidth}m{0.026\textwidth}}
    \hline
	Backbone & \multicolumn{3}{|c|}{DSH} & \multicolumn{3}{c|}{HashNet} & \multicolumn{3}{c|}{GreedyHash} & \multicolumn{3}{c|}{IDHN} & \multicolumn{3}{c|}{CSQ} & \multicolumn{3}{c}{DPN} \\ \cline{2-19}
	Network & 16b & 32b & 64b & 16b & 32b & 64b & 16b & 32b & 64b & 16b & 32b & 64b & 16b & 32b & 64b & 16b & 32b & 64b \\ \hline
	
	\multicolumn{19}{c}{Results on CIFAR-10 dataset (mAP@54000 in \%)}\\\hline
	AlexNet & 73.5 & 77.7 & 78.5 & 65.1 & 77.7 & 79.8 & 76.1 & 78.0 & 81.6 & 75.9 & 76.4 & 76.7 & 75.9 & 78.0 & 78.3 & 73.5 & 76.1 & 77.9 \\
	ResNet & 69.1 & 67.0 & 52.4 & 55.5 & 86.8 & 88.0 & 81.5 & 83.5 & 86.3 & 75.6 & 85.6 & 86.5 & 83.6 & 82.8 & 83.7 & 82.0 & 83.8 & 82.0 \\
	VTS32 & 93.0 & 94.6 & 95.3 & 96.2 & 97.1 & 95.6 & 94.8 & 95.6 & 94.3 & 95.9 & 95.8 & 96.0 & 96.0 & 96.1 & 95.2 & 95.1 & 95.6 & 95.0 \\
	VTS16 & \textbf{98.0} & \textbf{97.4} & \textbf{97.4} & \textbf{97.8} & \textbf{98.2} & \textbf{98.3} & \textbf{97.0} & \textbf{97.0} & \textbf{98.0} & \textbf{97.6} & \textbf{97.5} & \textbf{98.4} & \textbf{97.9} & \textbf{97.9} & \textbf{97.6} & \textbf{97.4} & \textbf{96.7} & \textbf{97.5} \\\hline
	
	\multicolumn{19}{c}{Results on CIFAR-10 dataset (mAP@All in \%)}\\\hline
	AlexNet & 76.0 & 78.5 & 79.0 & 65.9 & 78.0 & 80.1 & 76.9 & 81.8 & 81.1 & 76.4 & 77.5 & 78.1 & 79.6 & 78.3 & 76.8 & 76.2 & 79.8 & 79.2 \\
	ResNet & 64.3 & 57.5 & 56.3 & 58.9 & 86.1 & 85.7 & 82.2 & 85.8 & 87.0 & 85.5 & 85.9 & 85.6 & 84.9 & 83.5 & 85.1 & 83.7 & 83.3 & 84.0 \\
	VTS32 & 93.7 & 95.0 & 95.4 & 96.2 & 95.9 & 96.6 & 95.6 & 94.9 & 95.8 & 95.5 & 95.5 & 94.3 & 96.2 & 97.0 & 96.1 & 96.3 & 95.5 & 95.2 \\
	VTS16 & \textbf{97.4} & \textbf{98.1} & \textbf{98.3} & \textbf{98.1} & \textbf{97.9} & \textbf{98.2} & \textbf{97.0} & \textbf{97.7} & \textbf{97.5} & \textbf{97.8} & \textbf{98.0} & \textbf{97.9} & \textbf{97.6} & \textbf{97.7} & \textbf{97.6} & \textbf{96.8} & \textbf{96.5} & \textbf{97.7} \\\hline
	
	\multicolumn{19}{c}{Results on ImageNet dataset (mAP@1000 in \%)}\\\hline
	AlexNet & 50.4 & 60.6 & 64.4 & 35.9 & 57.2 & 64.8 & 57.7 & 65.8 & 68.2 & 02.0 & 37.7 & 49.9 & 60.5 & 67.5 & 69.6 & 59.2 & 67.8 & 69.2 \\
	ResNet & 47.3 & 67.3 & 75.3 & 20.5 & 47.8 & 66.0 & 81.2 & 84.6 & 85.6 & 04.3 & 36.5 & 58.0 & 83.6 & 86.0 & 86.6 & 83.3 & 85.7 & 86.6 \\
	VTS32 & 51.8 & 80.6 & 86.0 & 61.1 & 81.2 & 84.4 & 86.5 & 86.5 & 88.0 & 71.0 & \textbf{72.7} & \textbf{66.8} & 86.6 & 88.3 & 88.6 & 84.5 & 88.1 & 88.7 \\
	VTS16 & \textbf{85.6} & \textbf{90.8} & \textbf{91.7} & \textbf{81.7} & \textbf{88.3} & \textbf{89.5} & \textbf{90.3} & \textbf{91.1} & \textbf{91.6} & \textbf{80.5} & 67.2 & 64.2 & \textbf{90.6} & \textbf{91.5} & \textbf{91.8} & \textbf{90.2} & \textbf{91.4} & \textbf{91.8}\\\hline
	
	\multicolumn{19}{c}{Results on NUS-Wide dataset (mAP@5000 in \%)}\\\hline
	AlexNet & 77.5 & 79.2 & 80.8 & 75.2 & 81.6 & 84.5 & 74.8 & \textbf{78.8} & \textbf{80.0} & 79.1 & 80.6 & 81.3 & 78.0 & 82.1 & 83.4 & 75.0 & 81.3 & 83.5 \\
	ResNet & 77.1 & 79.7 & 79.1 & 76.8 & 82.3 & 83.9 & 75.6 & 77.7 & 78.8 & 80.9 & 79.9 & 79.4 & 79.3 & 83.6 & 83.8 & 78.9 & 82.5 & 84.0 \\
	VTS32 & 79.6 & 80.5 & \textbf{82.9} & 77.6 & 83.8 & 86.2 & 75.8 & 78.2 & 79.7 & 83.3 & 84.1 & 84.1 & \textbf{82.4} & \textbf{85.5} & \textbf{86.1} & \textbf{80.2} & \textbf{84.1} & \textbf{86.1} \\
	VTS16 & \textbf{79.8} & \textbf{81.7} & \textbf{82.9} & \textbf{79.2} & \textbf{85.2} & \textbf{87.3} & \textbf{76.2} & 76.9 & 78.6 & \textbf{84.1} & \textbf{85.2} & \textbf{85.1} & 81.9 & 84.6 & 85.3 & 79.7 & 83.5 & 84.6 \\\hline
	
	\multicolumn{19}{c}{Results on MS-COCO dataset (mAP@5000 in \%)}\\\hline
	AlexNet & 63.5 & 65.2 & 67.5 & 63.3 & 68.9 & 73.3 & 63.4 & 69.7 & 72.7 & 65.6 & 68.7 & 69.9 & 63.0 & 72.0 & 76.6 & 63.3 & 72.3 & 76.8 \\
	ResNet & 69.0 & 72.3 & 75.5 & 64.4 & 68.1 & 72.8 & 69.5 & 75.1 & 77.7 & 71.3 & 68.6 & 58.6 & 71.2 & 81.8 & 87.2 & 71.8 & 80.9 & 85.2 \\
	VTS32 & 71.9 & 75.8 & 76.9 & 70.6 & 76.8 & 80.6 & 75.5 & 79.6 & 80.3 & 76.3 & 79.9 & 78.4 & 77.0 & 86.5 & 88.9 & 75.3 & 84.5 & 88.0 \\
	VTS16 & \textbf{80.5} & \textbf{84.3} & \textbf{85.0} & \textbf{75.2} & \textbf{82.8} & \textbf{86.5} & \textbf{78.0} & \textbf{81.2} & \textbf{82.3} & \textbf{77.8} & \textbf{82.8} & \textbf{83.0} & \textbf{77.9} & \textbf{87.4} & \textbf{91.1} & \textbf{78.5} & \textbf{85.8} & \textbf{89.7} \\\hline
    \end{tabular}
    \label{table:retrieval_results}
\end{table*}

\section{Experimental Setup}
This section describes the experimental setup, including datasets used for the retrieval experiments, evaluation metrics, network and training settings.

\subsection{Datasets Used}
We use four datasets which are widely adapted for image retrieval, including CIFAR10, ImageNet, NUS-Wide and MS-COCO. 
The CIFAR10 dataset \cite{cifar} is a collection of $60,000$ images from $10$ categories with $6,000$ images per category. We follow the standard experimental practice on the CIFAR10 dataset. Similar to \cite{dhn}, \cite{dch}, \cite{transhash}, we randomly sample $5,000$ images for the training set with $500$ images from each category. Then, the query set having $1000$ images is sampled randomly with $100$ images per category. The rest of the images are used as the database. This is referred to as CIFAR10@54000. We also compute the results on the CIFAR10 dataset by following the protocol of \cite{dpn} which considers the training set also as part of the database. This is referred to as CIFAR10@All. 
ImageNet dataset is a subset of the Large Scale Visual Recognition Challenge (ILSVRC 2015) \cite{imagenet}. We use the standard retrieval protocol \cite{csq}, \cite{dpn} on the ImageNet dataset. Basically, $100$ classes are randomly sampled. All the validation set images of these classes are used in the query set. However, all the training set images of these classes are used as the database. From the database images, $13,000$ images are sampled as the training set.
NUS-Wide dataset is one of the widely used dataset for image retrieval \cite{nus-wide}. We follow the protocol used in \cite{dpn} where $21$ most frequent concepts are considered as image annotations leading to $195K$ images. The query set is sampled randomly with $100$ images per concept. The remaining images are considered as a database for retrieval. The training set consists of $500$ images per concept randomly sampled from the database.
MS-COCO dataset consists of images from $80$ categories \cite{mscoco}. We use the existing setup \cite{dch}, \cite{csq} of MS-COCO dataset for image retrieval which includes $5,000$, $10,000$ and $1,17,218$ images in query, training and retrieval sets, respectively.

\subsection{Evaluation Metrics}
We follow the standard practice \cite{csq}, \cite{dpn}, \cite{transhash} of evaluating the image retrieval methods and compute the Mean Average Precision (mAP). The mAP is computed for $54,000$, $59,000$, $1,000$, $5,000$ and $5,000$ retrieved images on CIFAR10@54000, CIFAR10@All, ImageNet, NUS-Wide and MS-COCO datasets, respectively.
Precision vs Recall curve is also used to observe the ROC characteristics.

\subsection{Network and Training Settings}
In the experiments, all the images are resized with $m = 224$. Two variants of the proposed Vision Transformer Hashing (VTS) are used for experiments, i.e., VTS32 and VTS16 having patch size $k = 32$ and $16$, respectively. Thus, the number of patches $N = 49$ and $196$ for VTS32 and VTS16, respectively. The hidden size ($de$) is set as $768$ in the experiments. The number of transformer blocks ($L$) as well as the number of attention heads ($A_h$) are $12$ in the VTS models. The hash code is generated with $16$, $32$ and $64$ bit length. All the models are trained for $150$ epochs with a batch size of $32$ using Adam optimizer with a learning rate of $1e^{-5}$. The testing is performed at every $30$ epochs and the best results are reported.

\begin{figure*}[!t]
\centering
\resizebox{1.01\linewidth}{!}{
\begin{tabular}{cccccc}
\includegraphics[trim={7 5 52 77},clip,width=\linewidth]{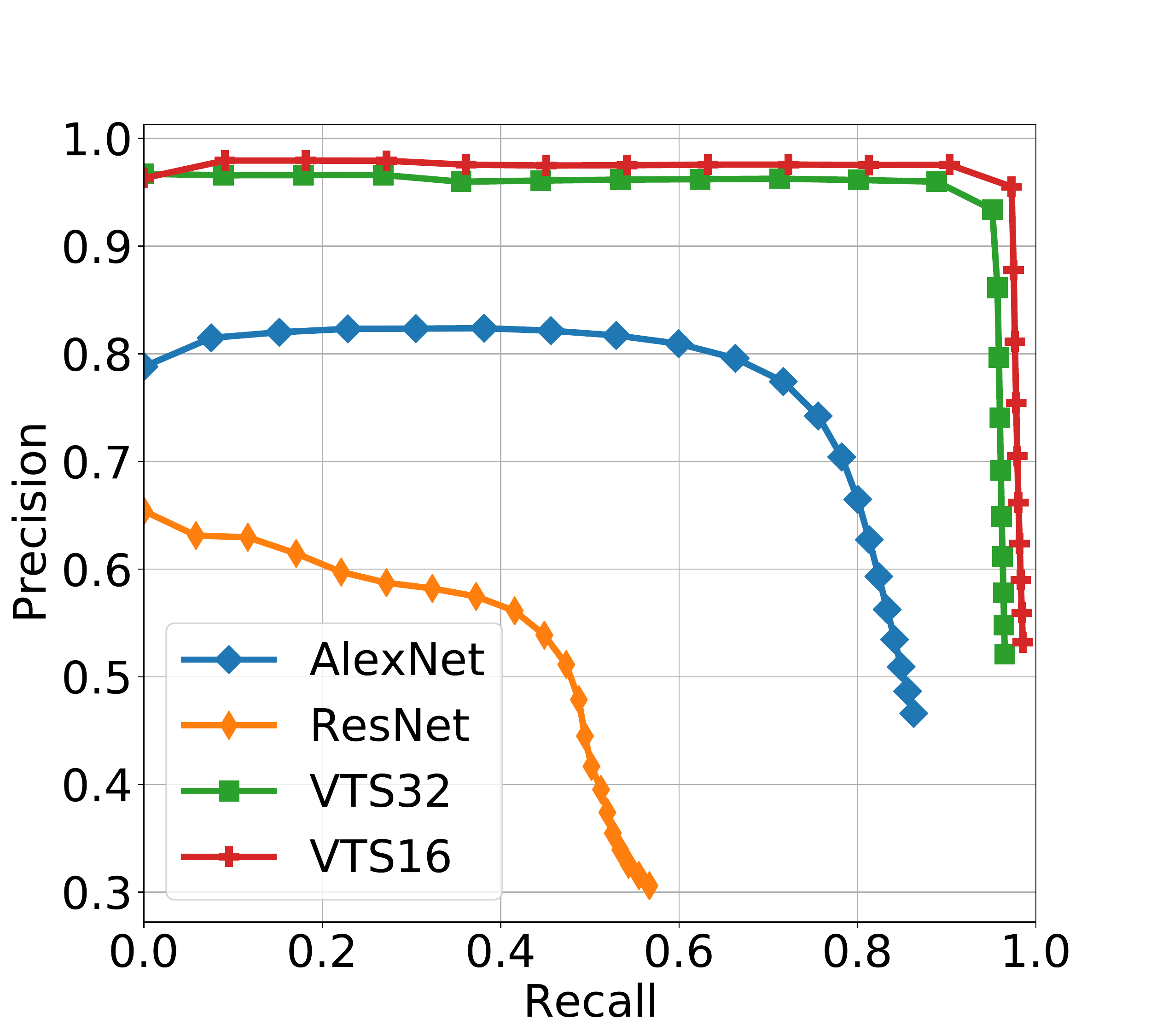} &
\includegraphics[trim={7 5 52 77},clip,width=\linewidth]{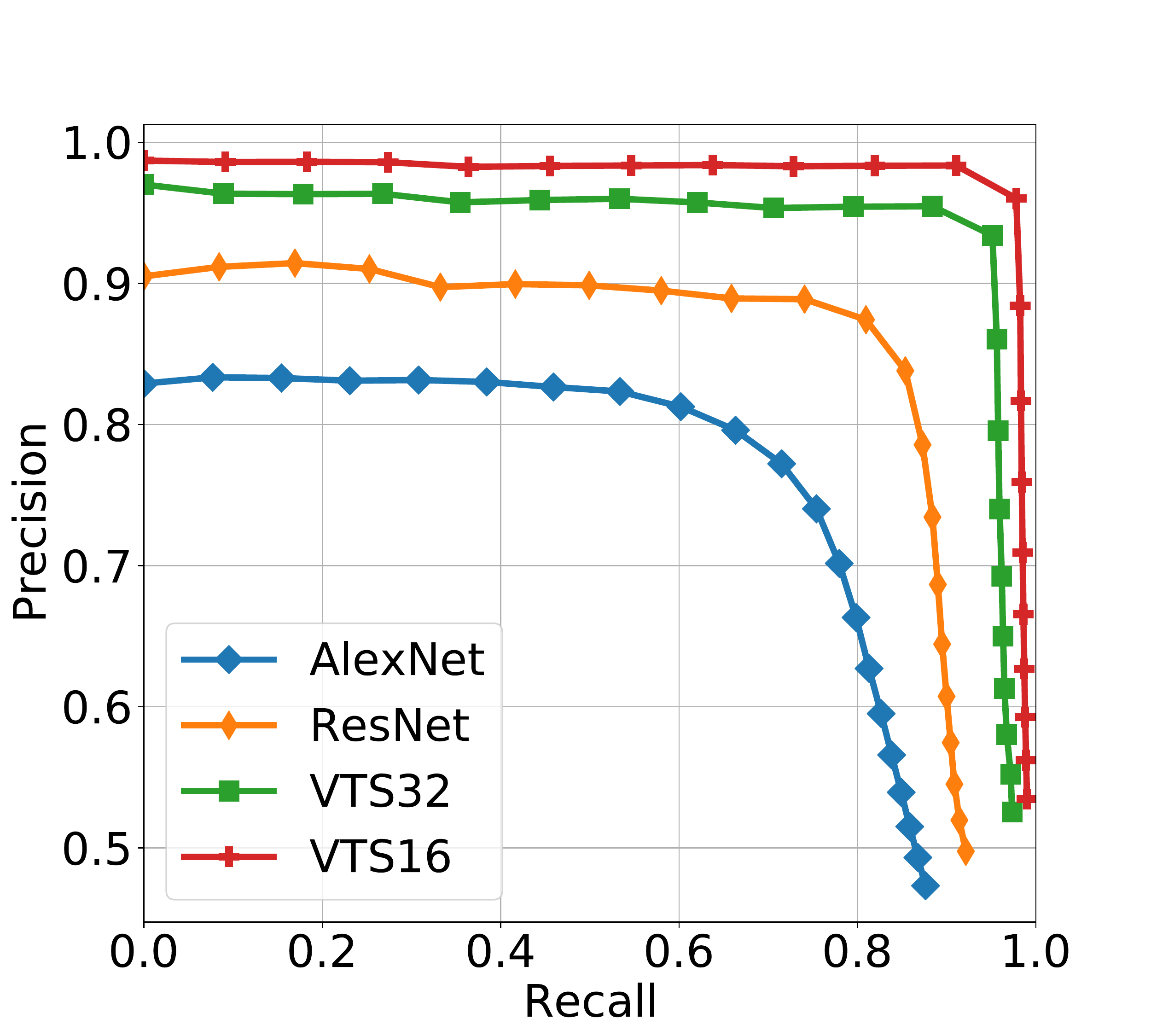}  &
\includegraphics[trim={7 5 52 77},clip,width=\linewidth]{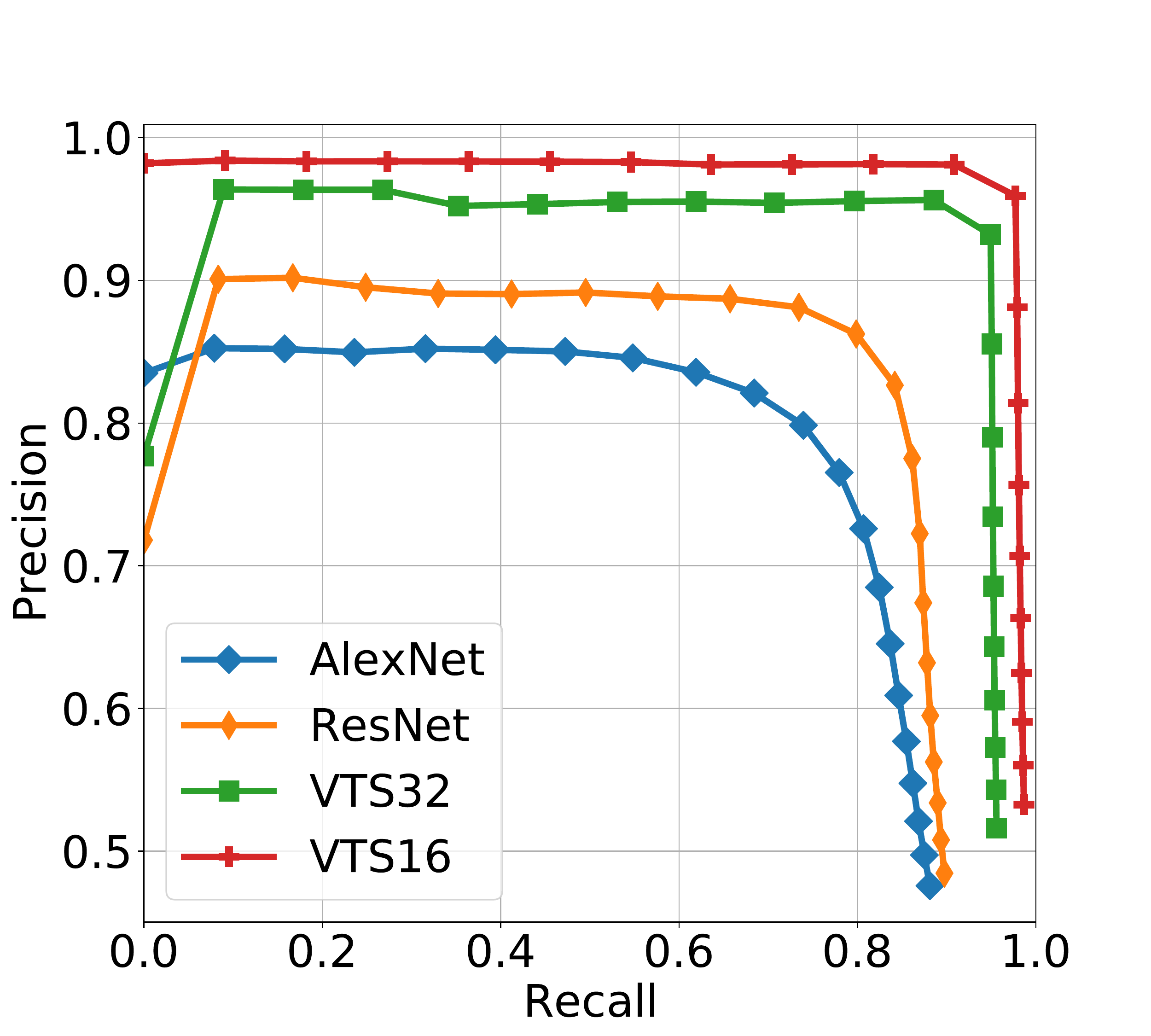}  &
\includegraphics[trim={7 5 52 77},clip,width=\linewidth]{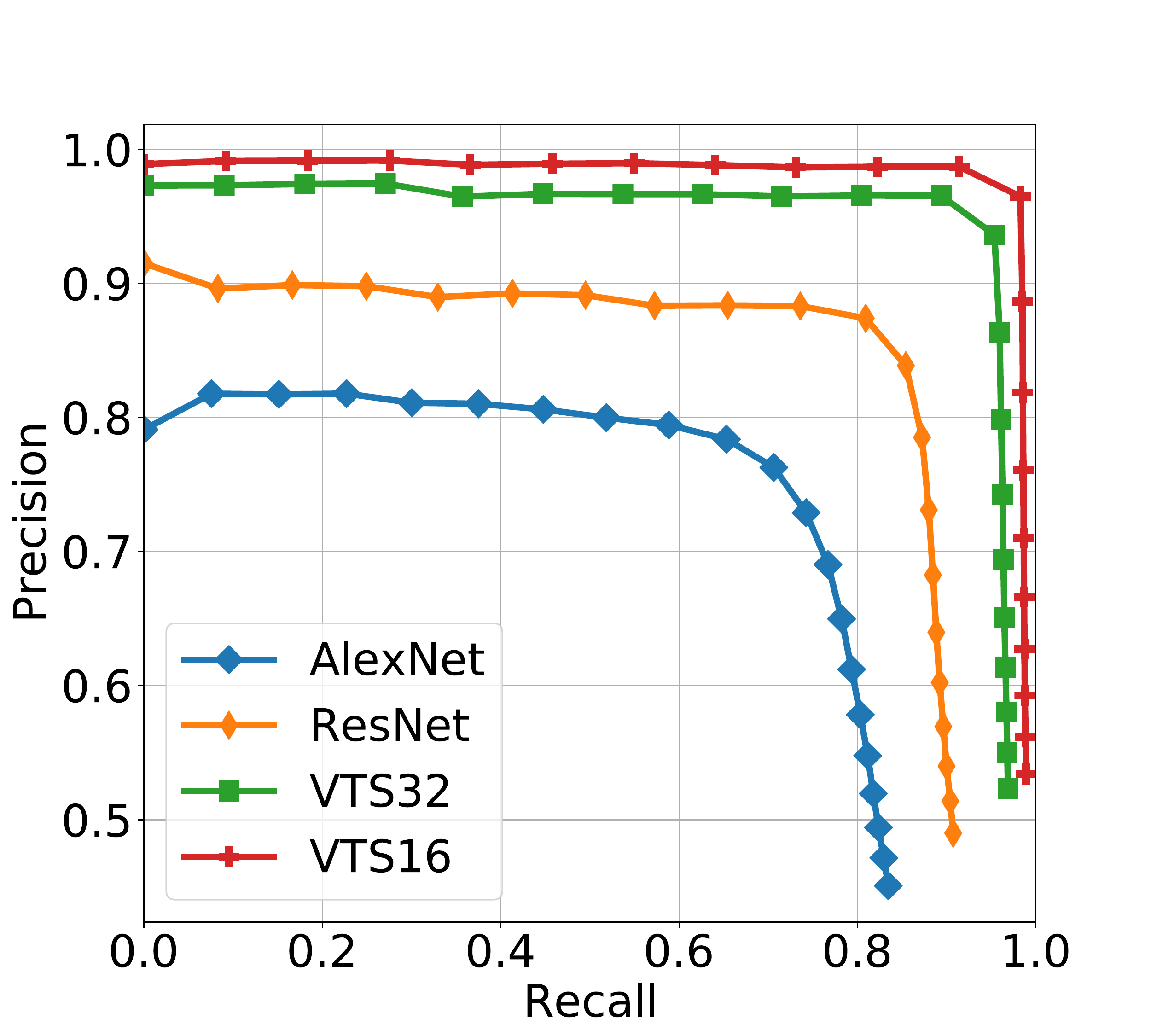}  &
\includegraphics[trim={7 5 52 77},clip,width=\linewidth]{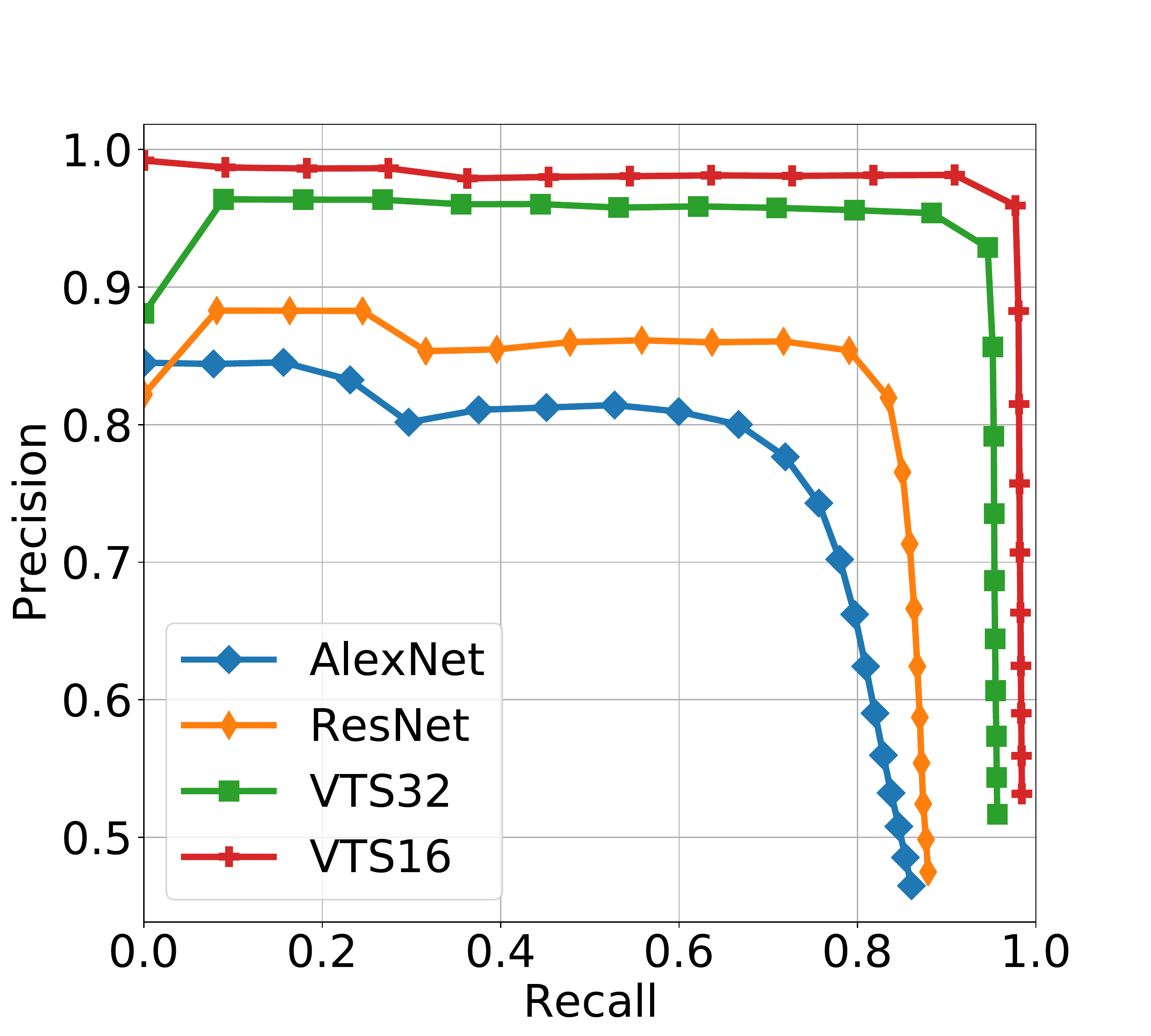}  &
\includegraphics[trim={7 5 52 77},clip,width=\linewidth]{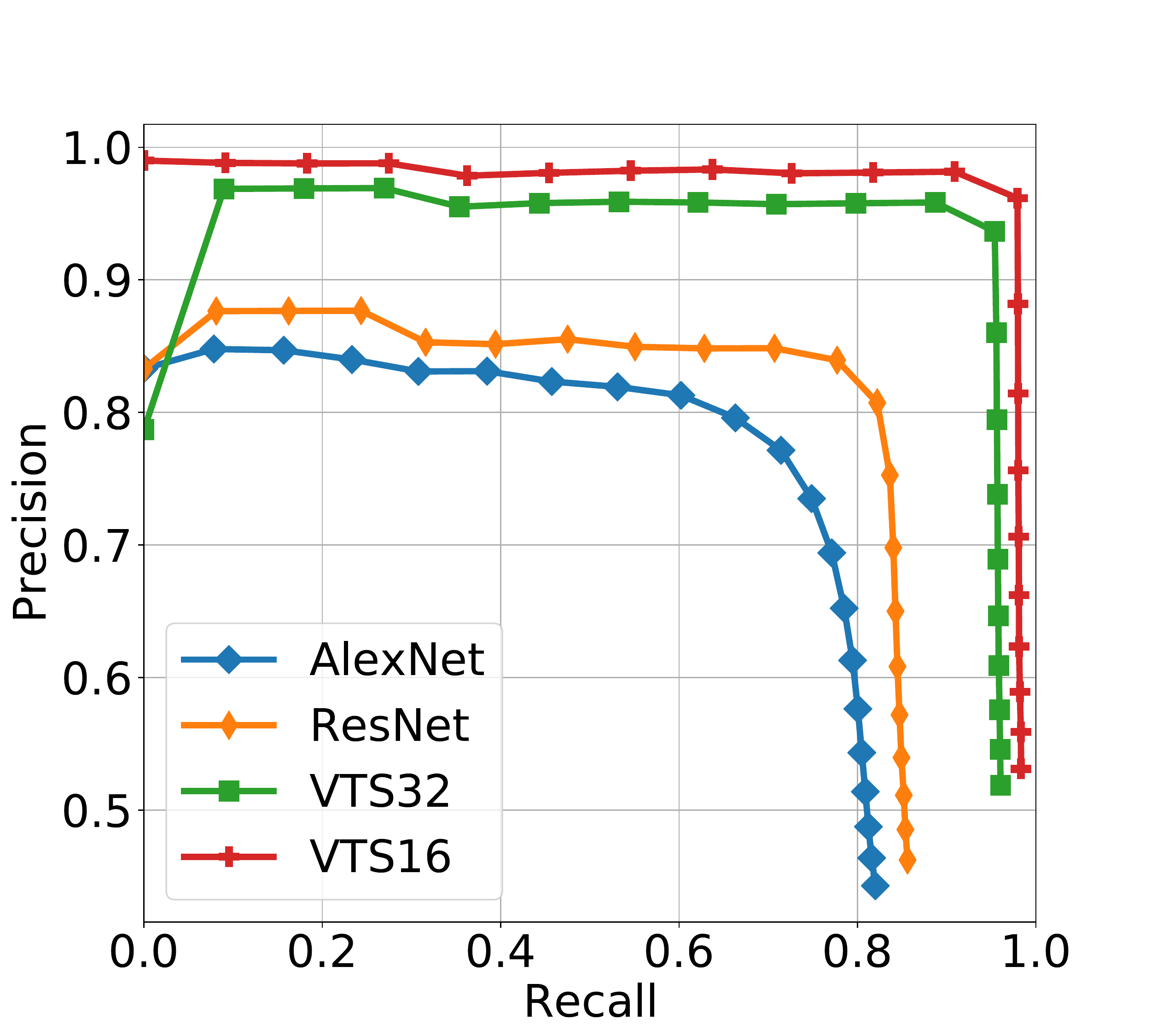} \\

\includegraphics[trim={7 5 52 77},clip,width=\linewidth]{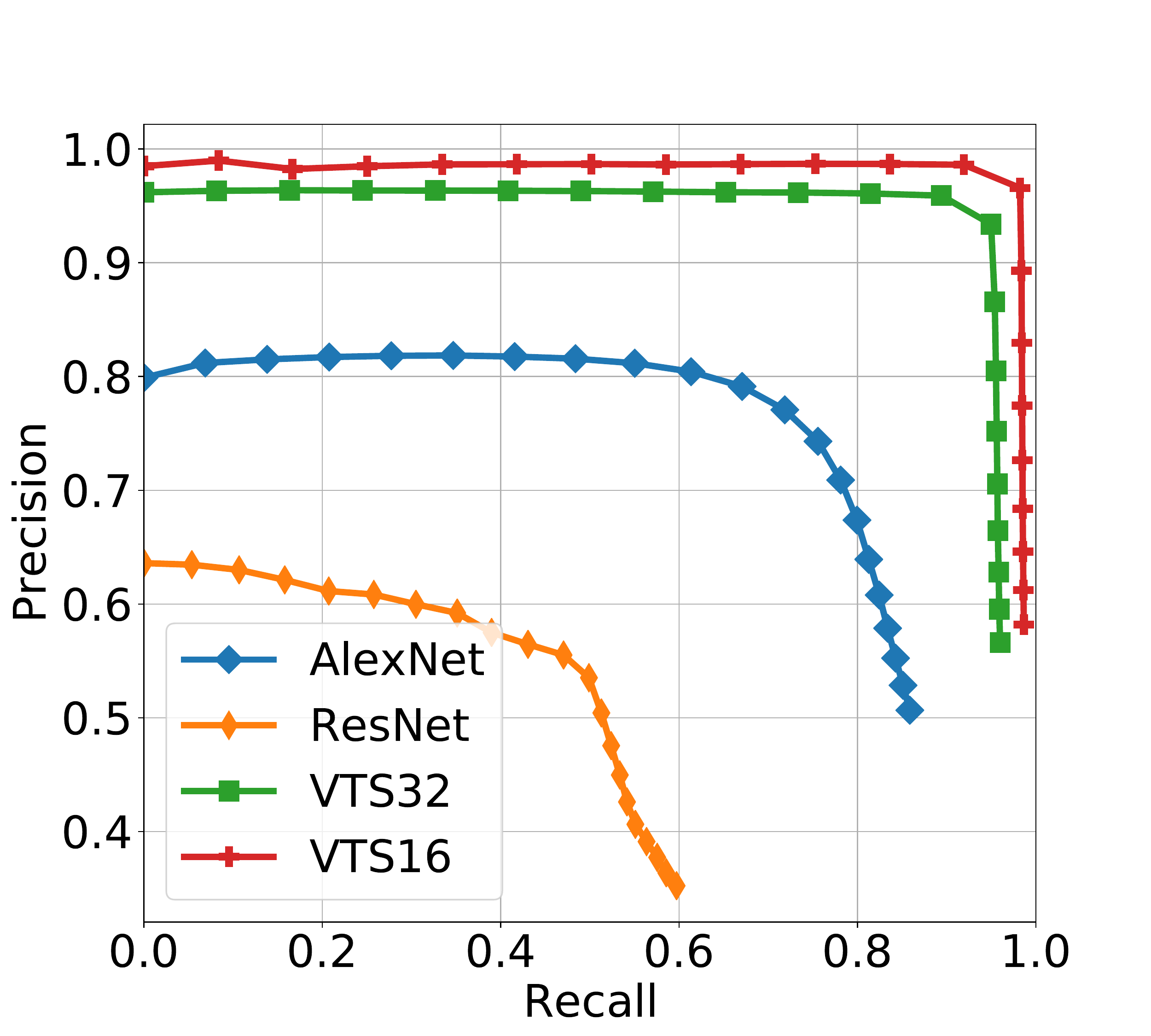} &
\includegraphics[trim={7 5 52 77},clip,width=\linewidth]{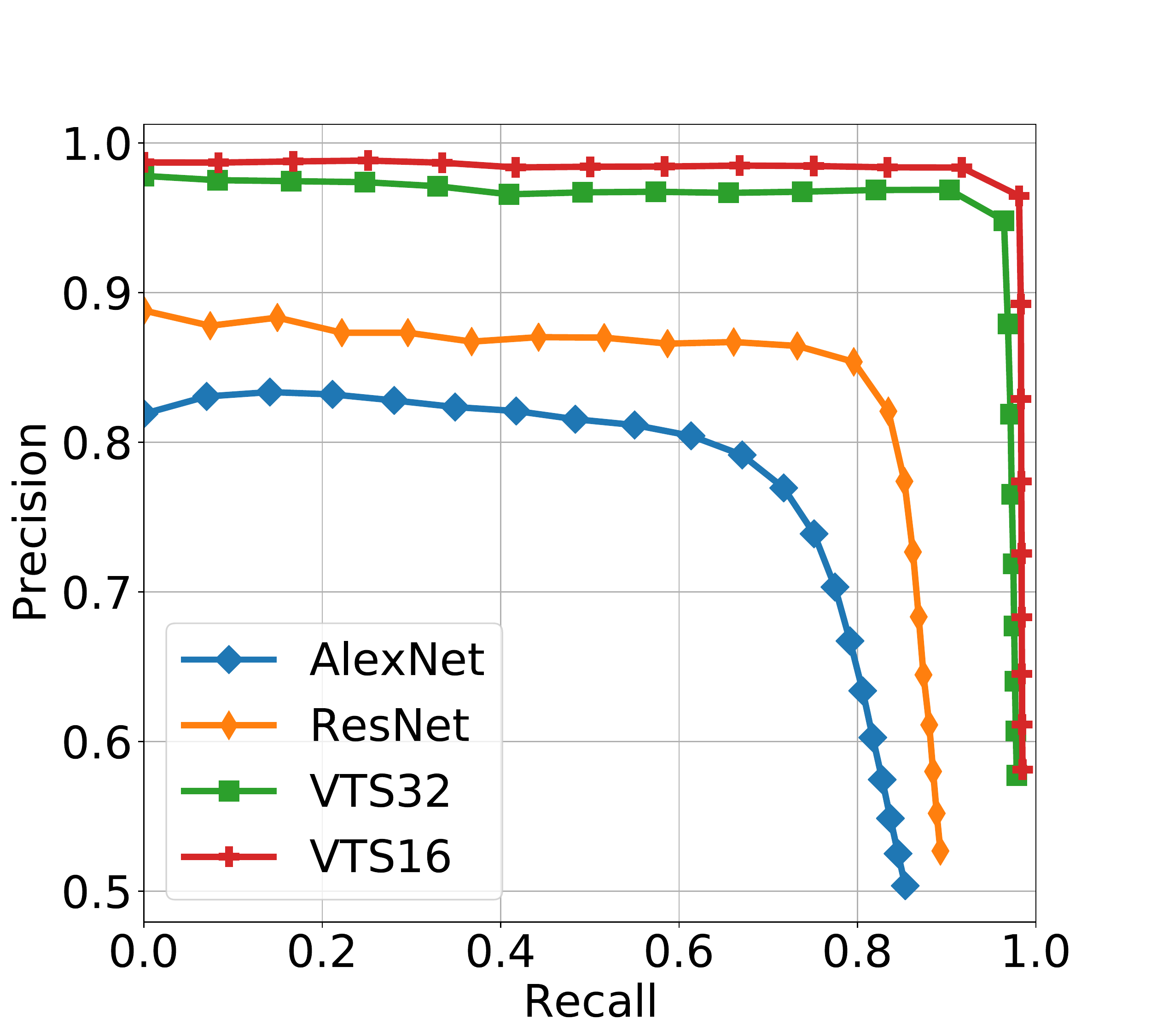}  &
\includegraphics[trim={7 5 52 77},clip,width=\linewidth]{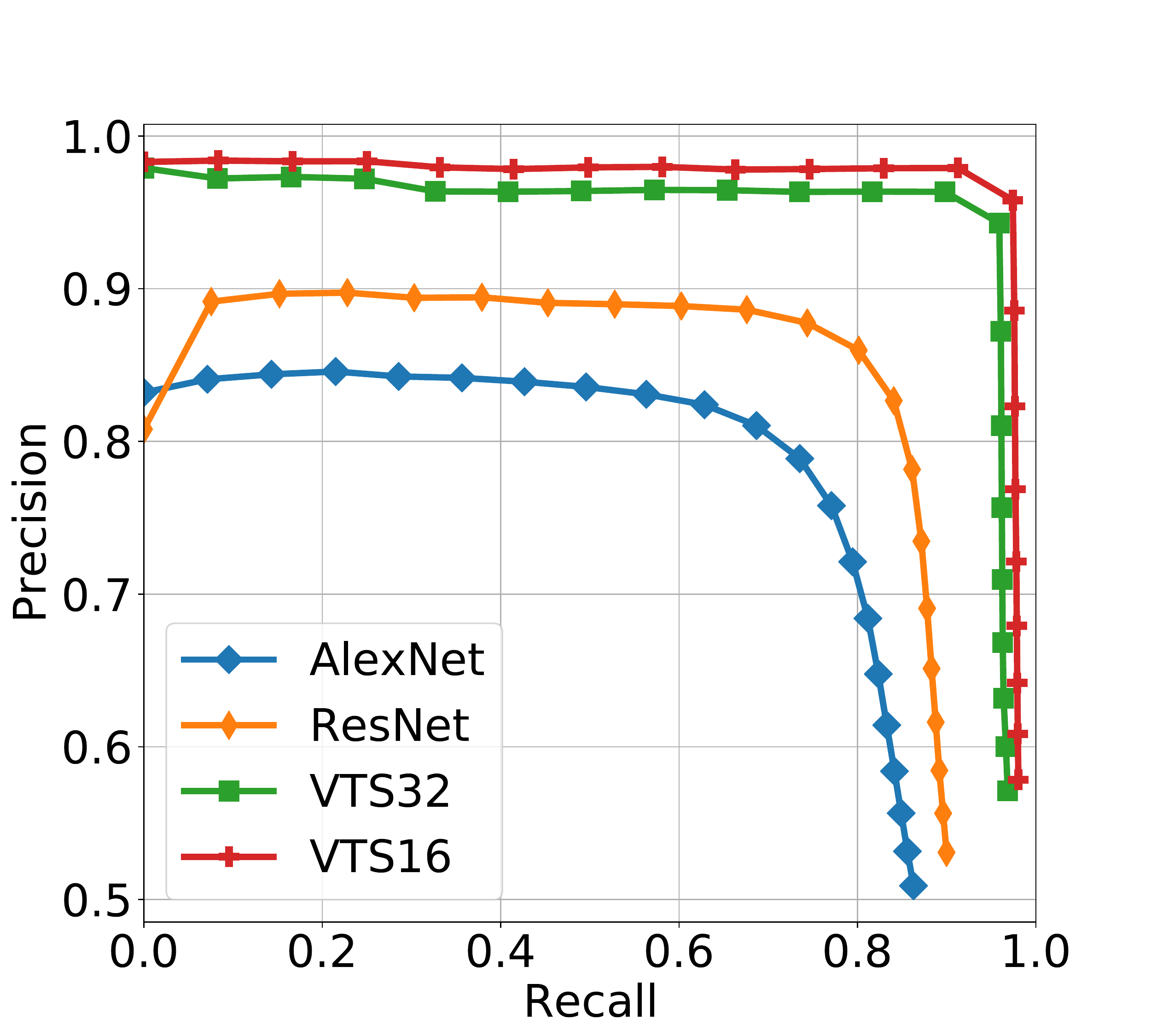}  &
\includegraphics[trim={7 5 52 77},clip,width=\linewidth]{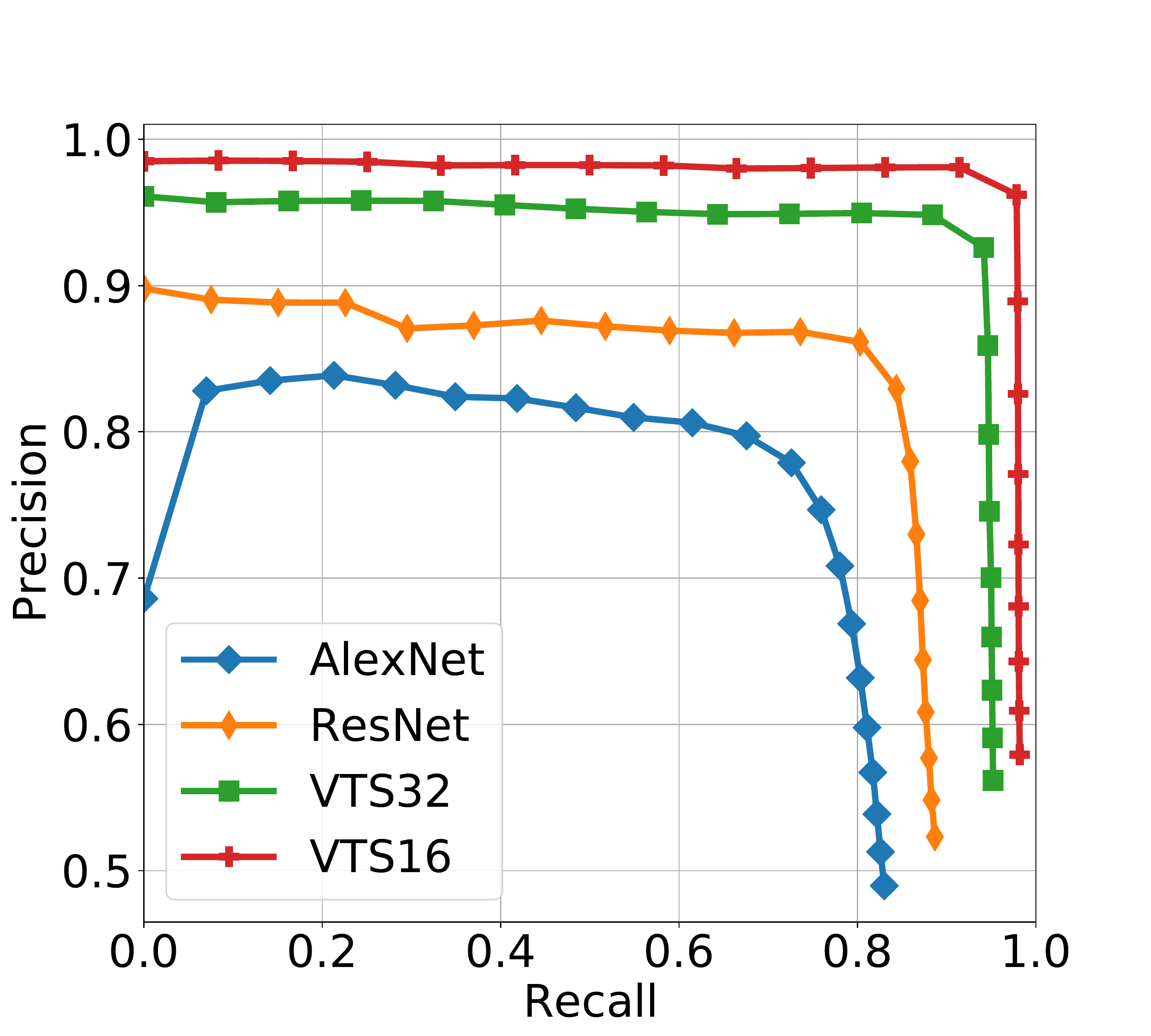}  &
\includegraphics[trim={7 5 52 77},clip,width=\linewidth]{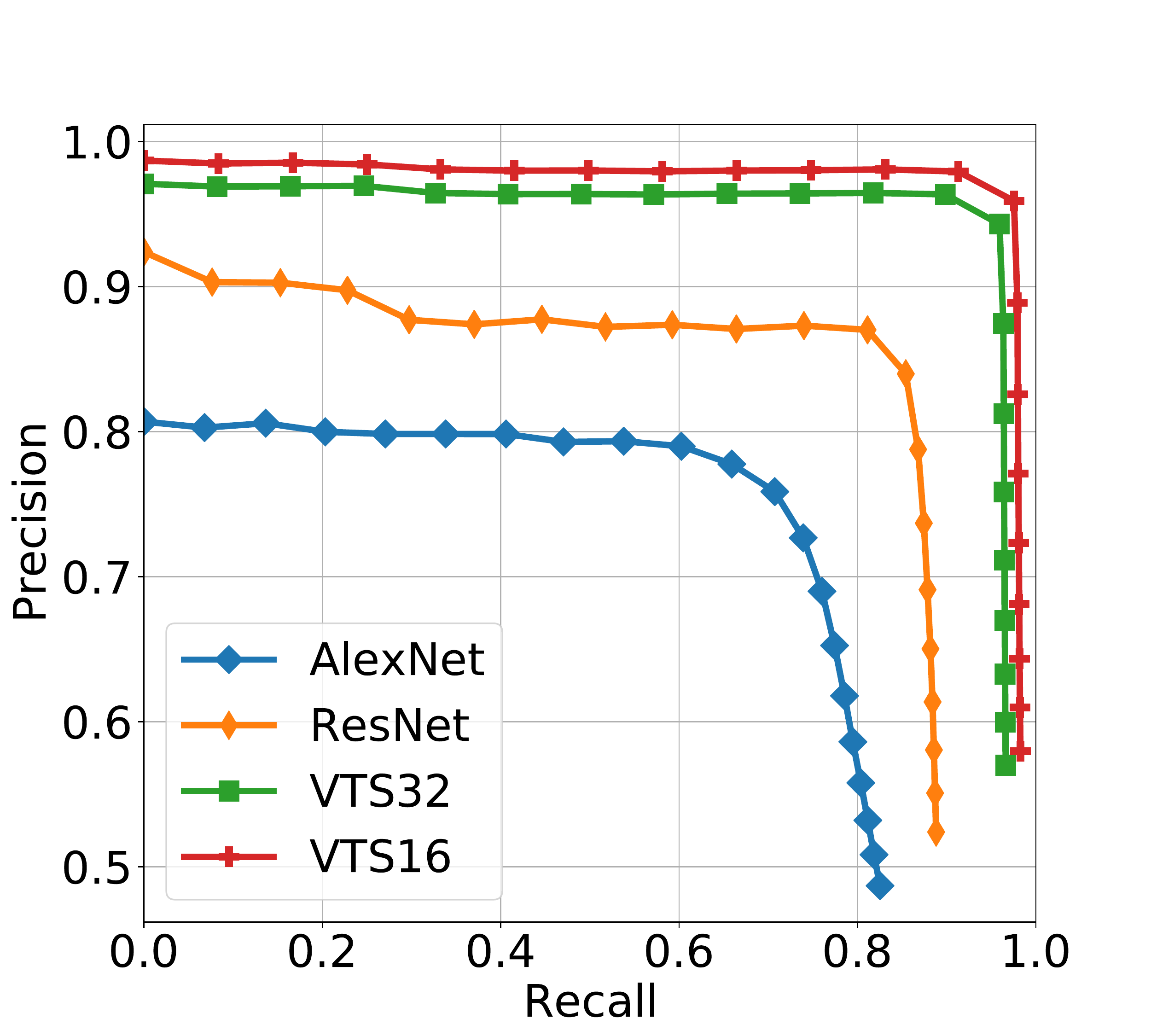}  &
\includegraphics[trim={7 5 52 77},clip,width=\linewidth]{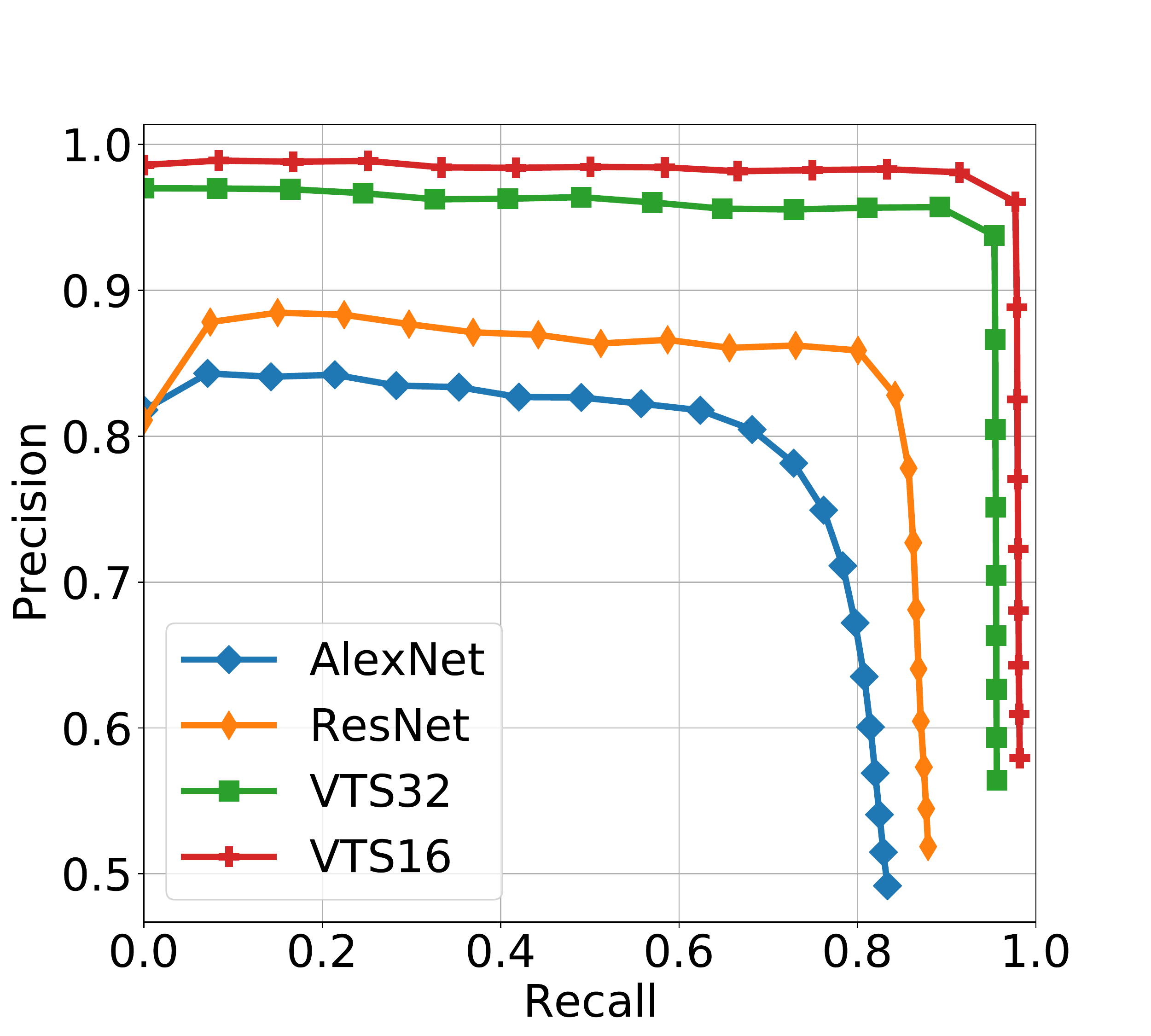} \\

\includegraphics[trim={15 5 65 77},clip,width=\linewidth]{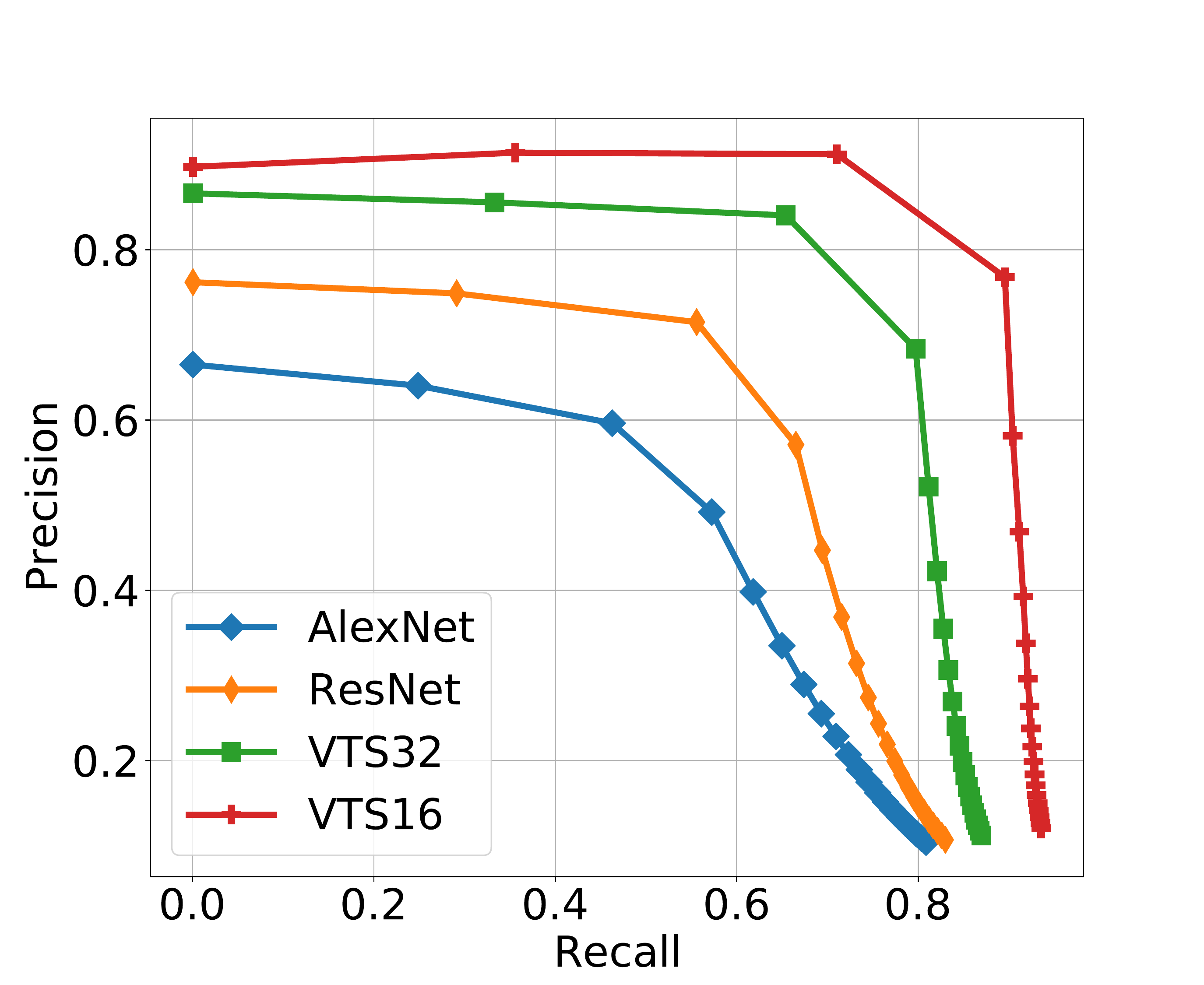} &
\includegraphics[trim={15 5 65 77},clip,width=\linewidth]{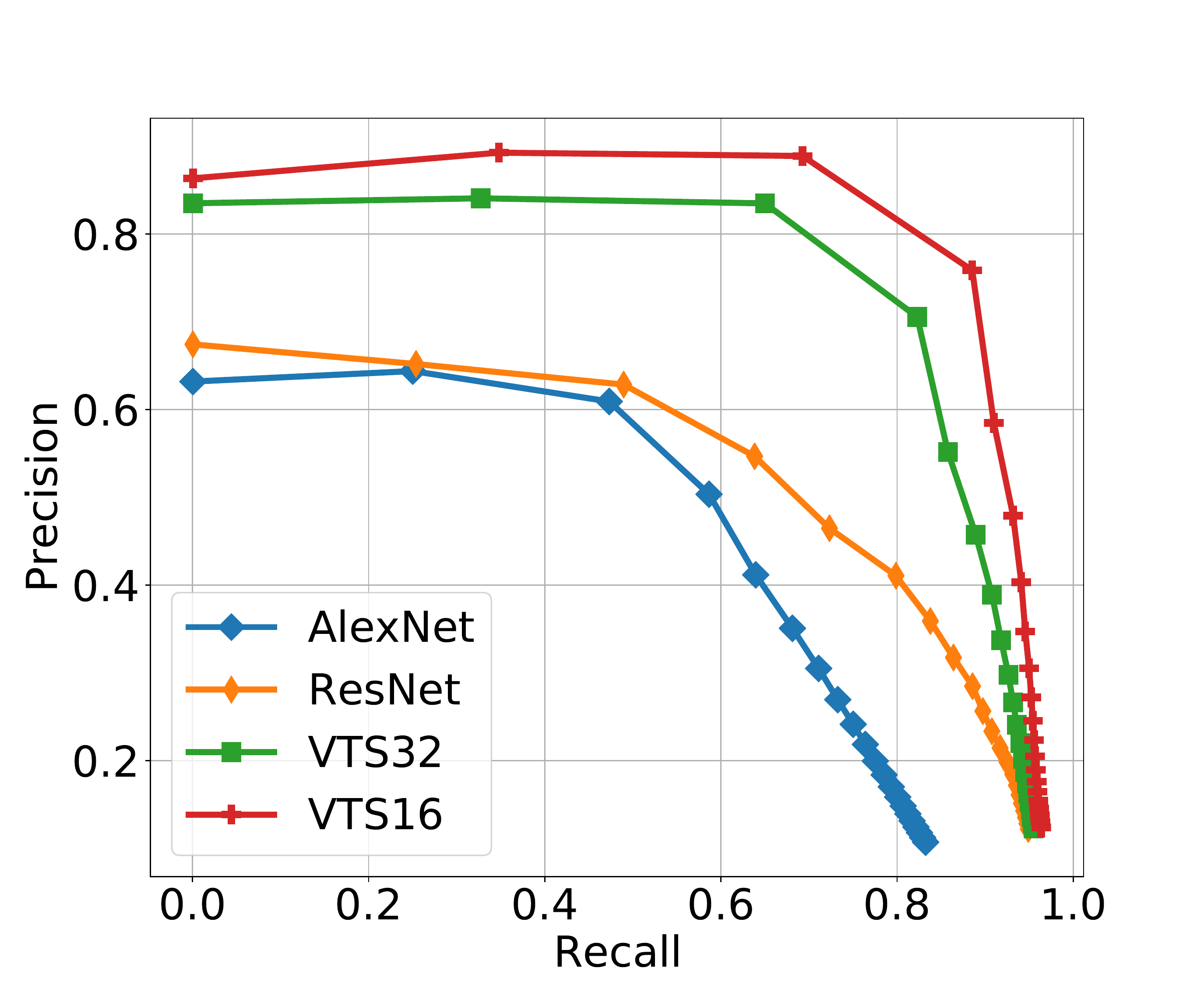}  &
\includegraphics[trim={15 5 65 77},clip,width=\linewidth]{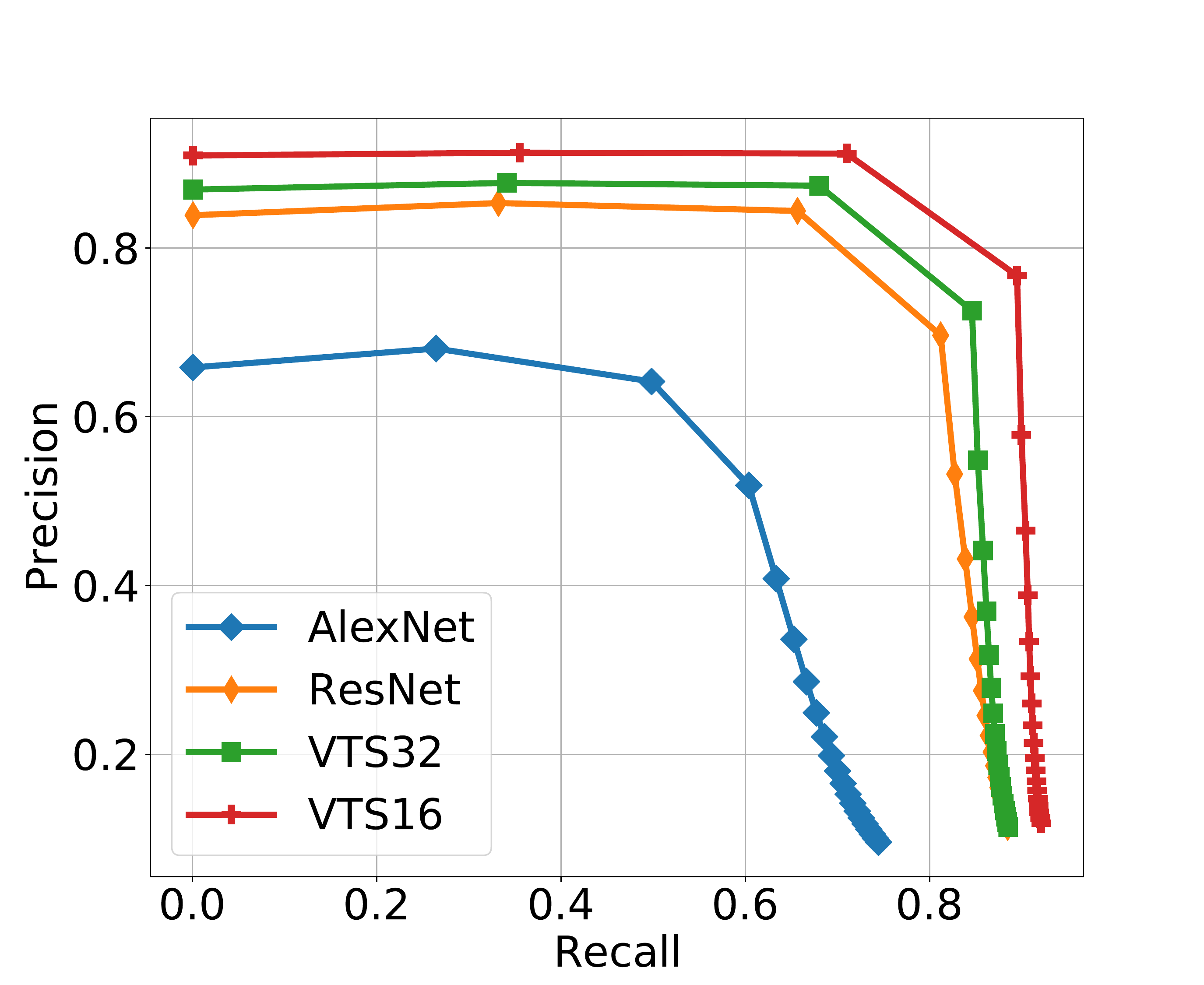}  &
\includegraphics[trim={15 5 65 77},clip,width=\linewidth]{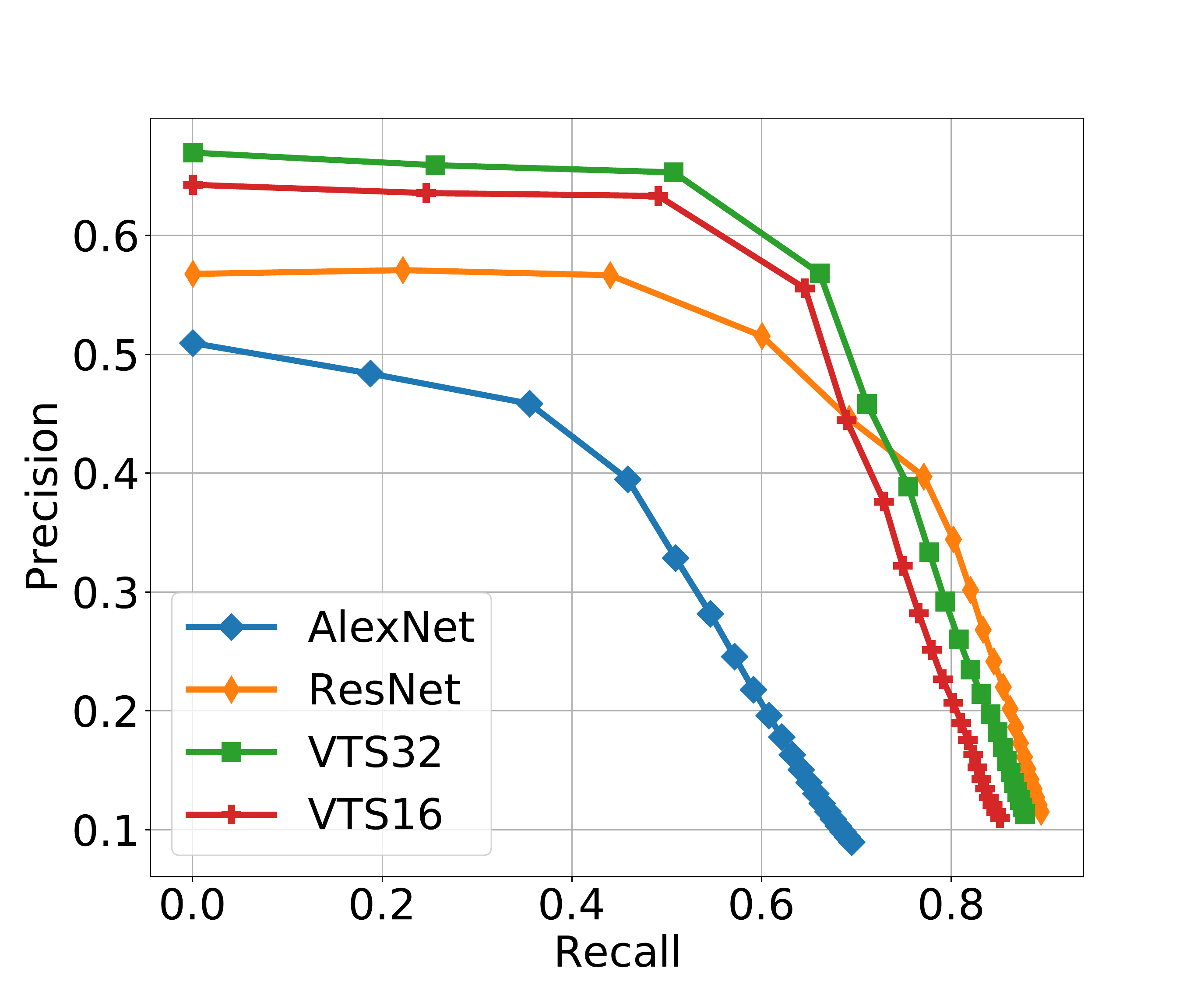}  &
\includegraphics[trim={15 5 65 77},clip,width=\linewidth]{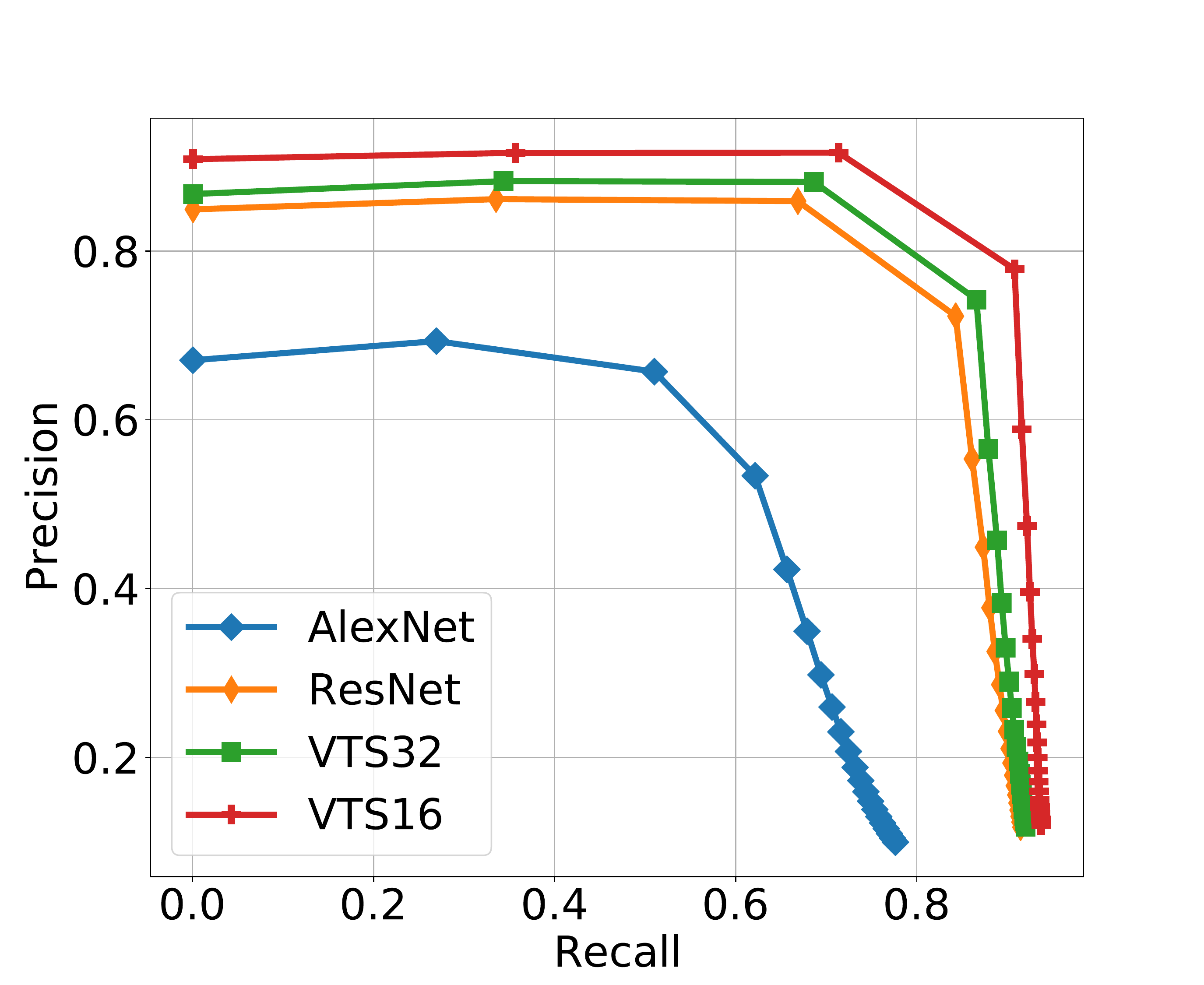}  &
\includegraphics[trim={15 5 65 77},clip,width=\linewidth]{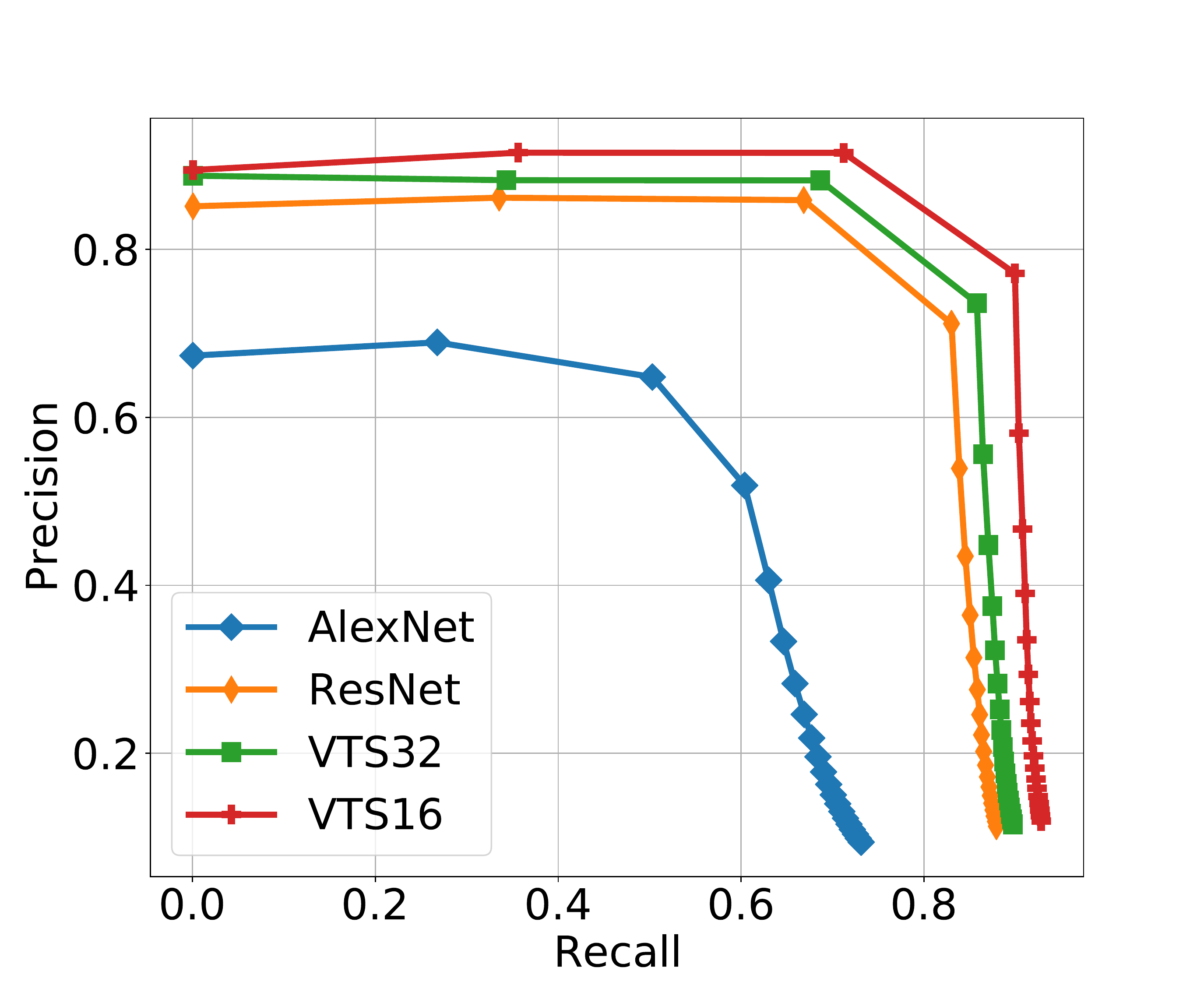} \\

\includegraphics[trim={0 5 65 77},clip,width=\linewidth]{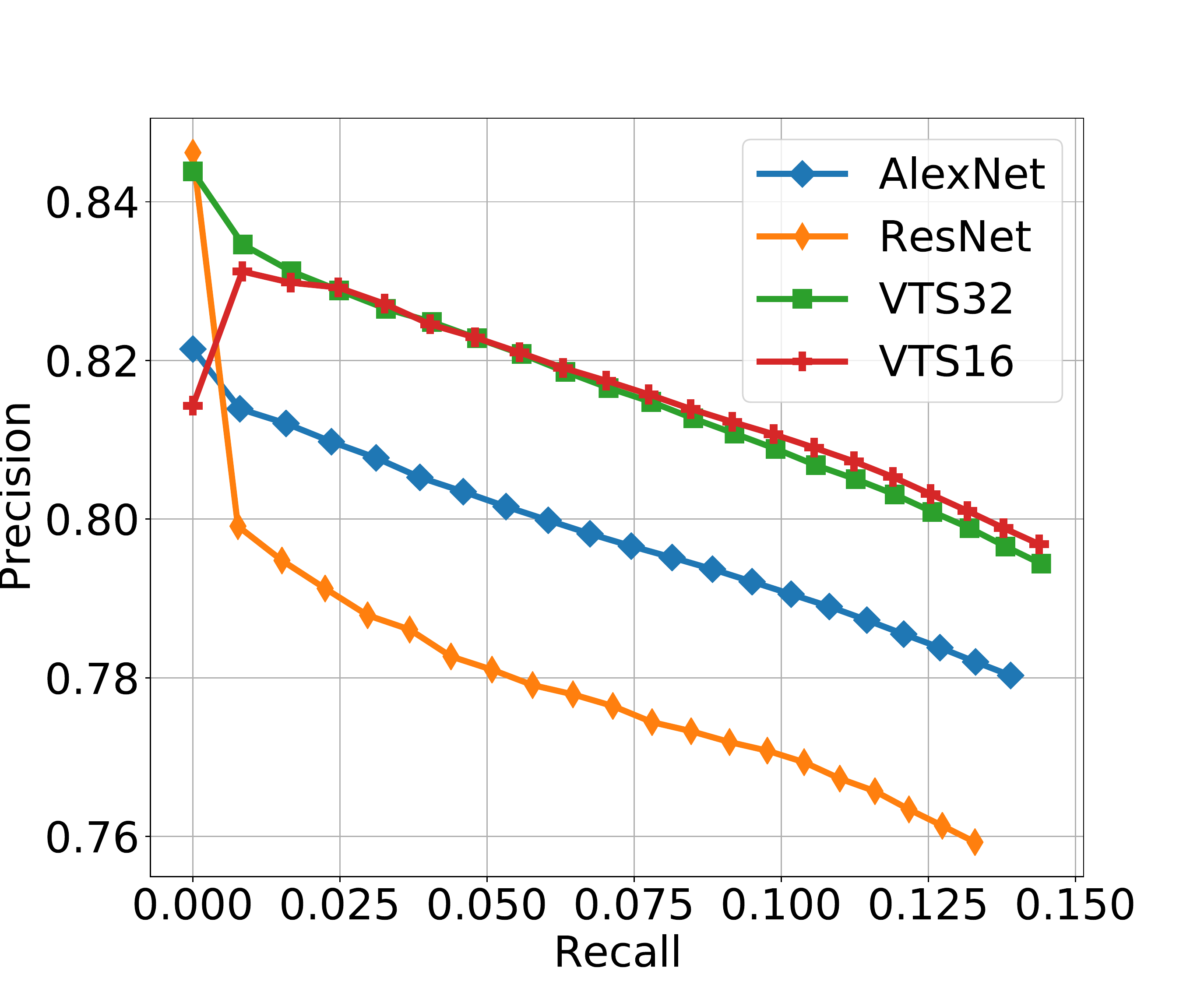} &
\includegraphics[trim={0 5 65 77},clip,width=\linewidth]{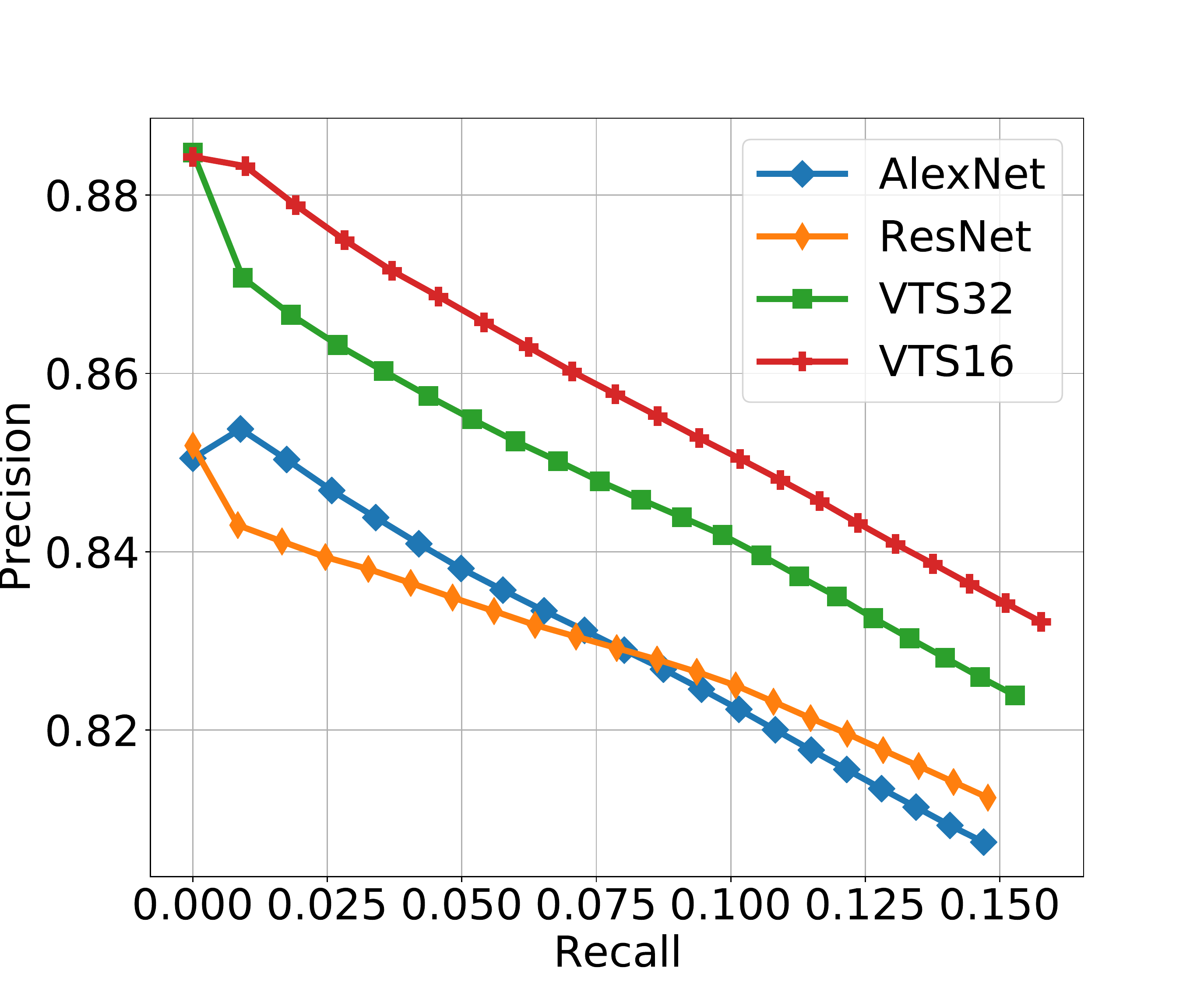}  &
\includegraphics[trim={0 5 65 77},clip,width=\linewidth]{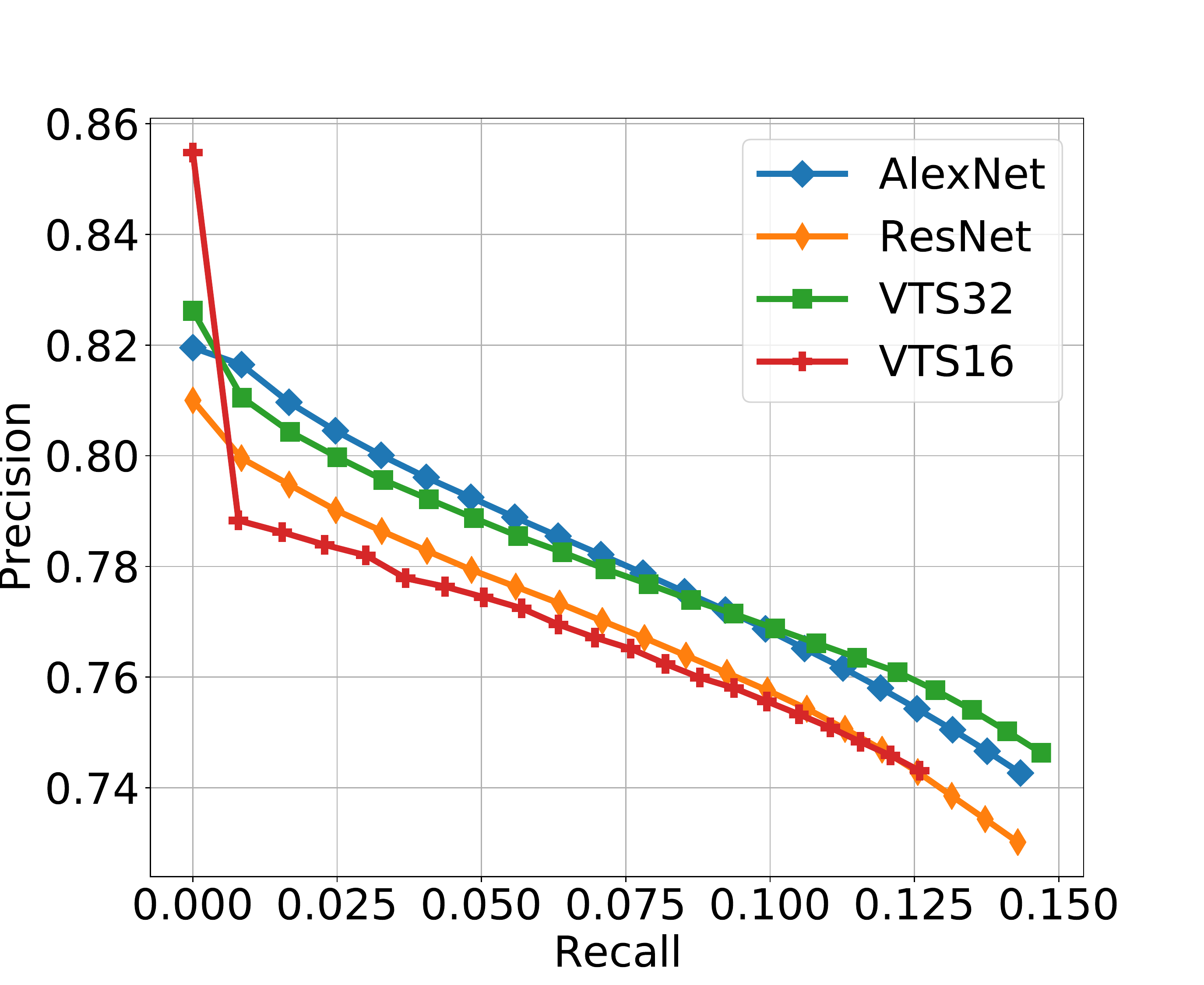}  &
\includegraphics[trim={0 5 65 77},clip,width=\linewidth]{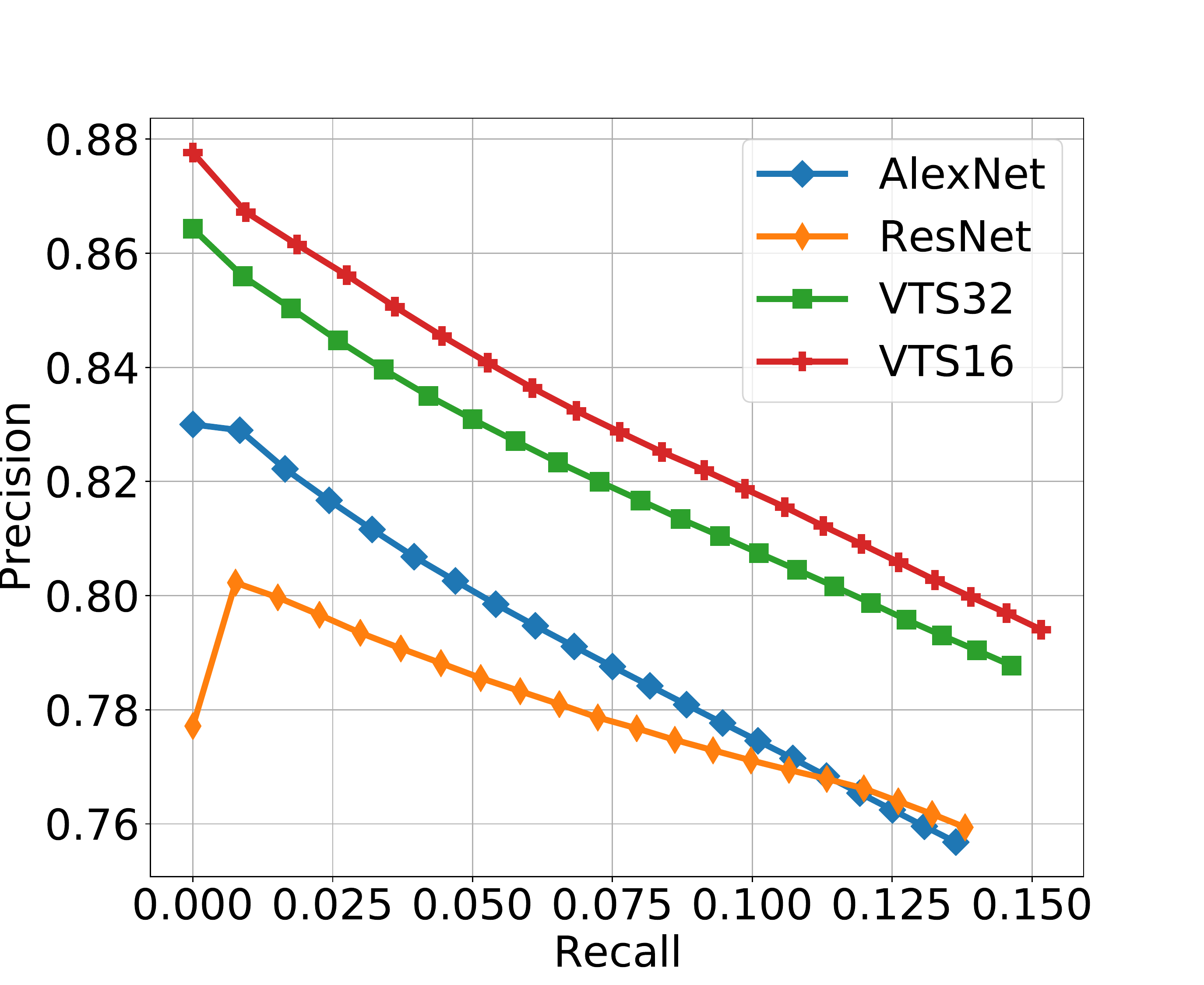}  &
\includegraphics[trim={0 5 65 77},clip,width=\linewidth]{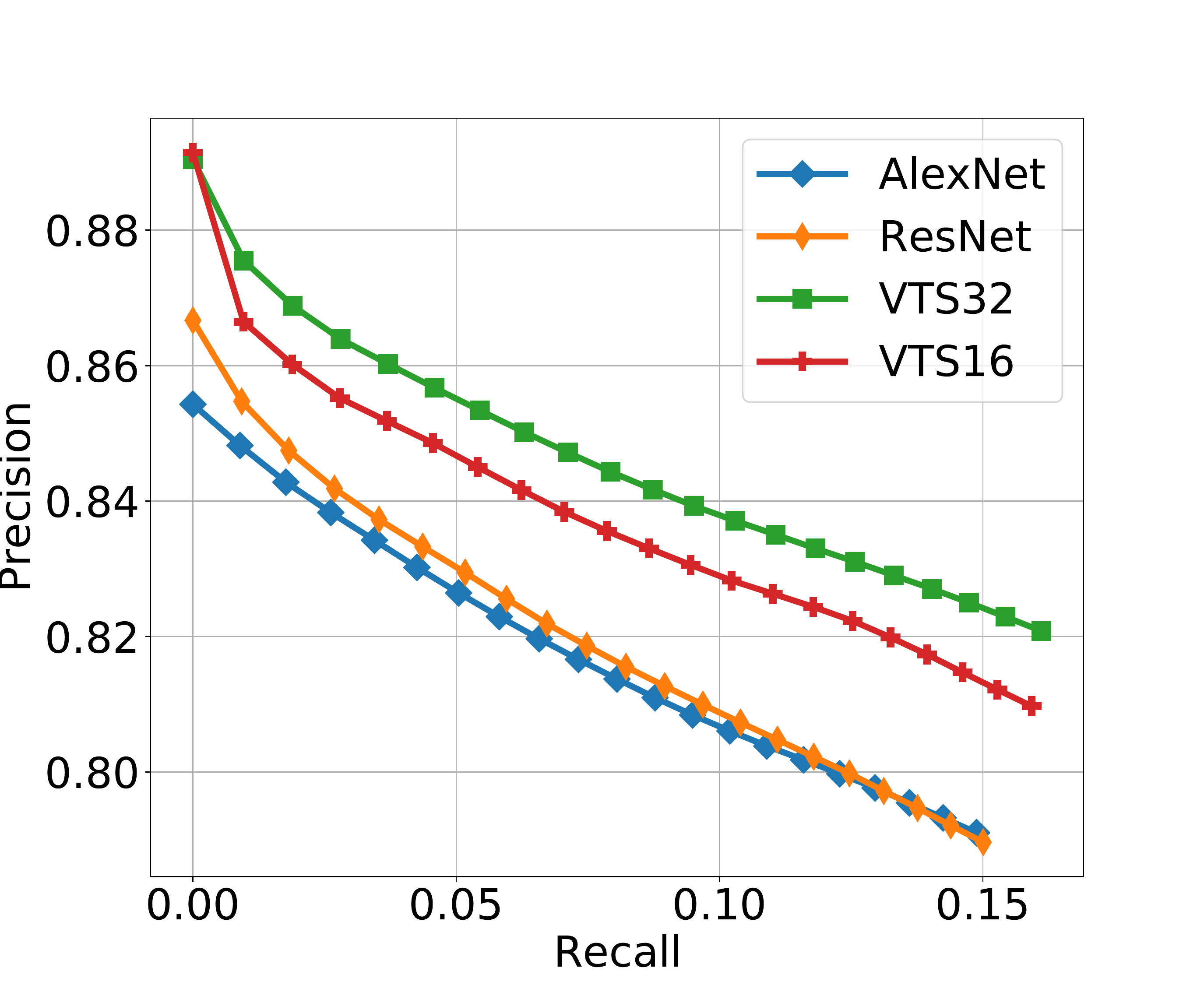}  &
\includegraphics[trim={0 5 65 77},clip,width=\linewidth]{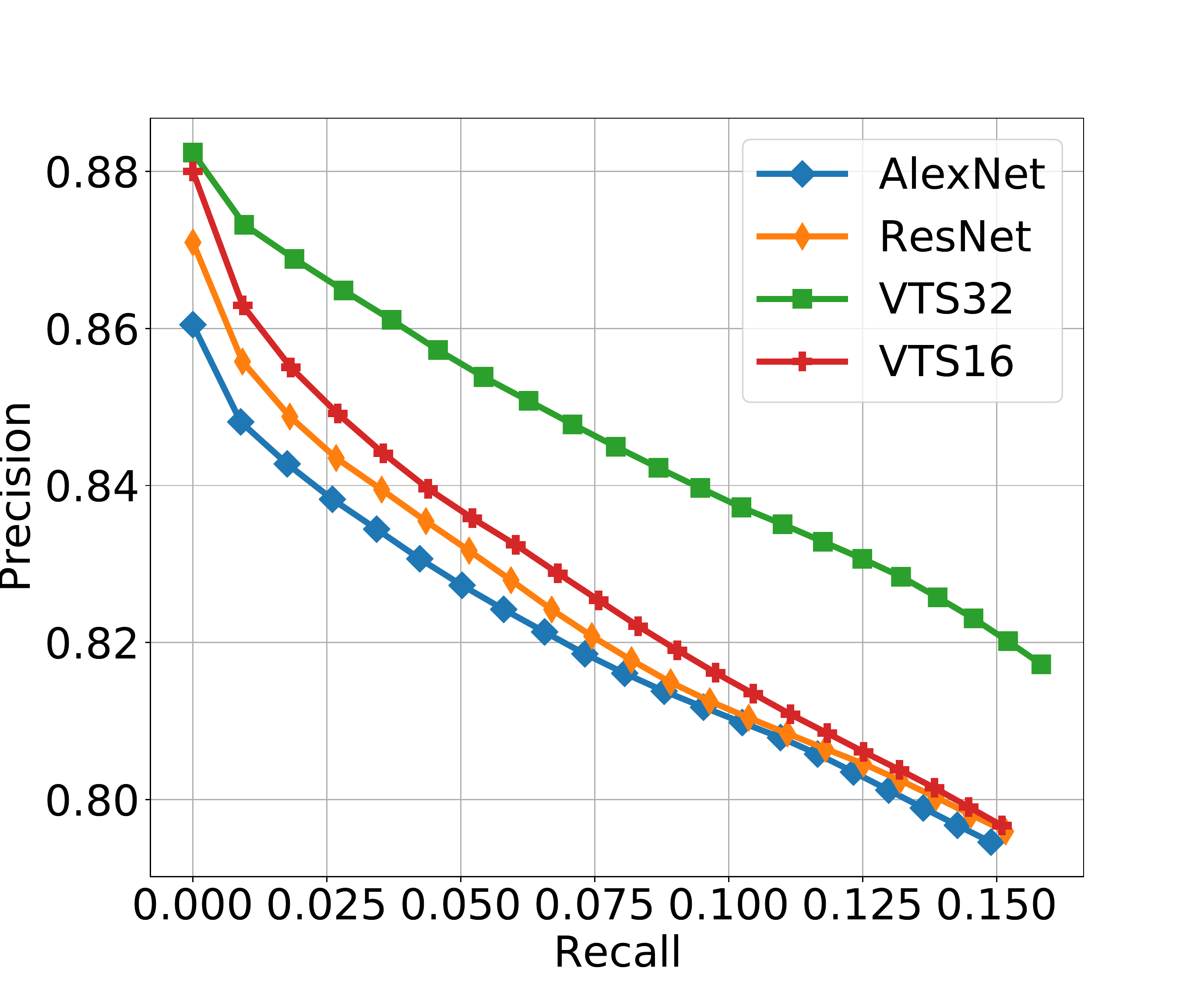} \\

\includegraphics[trim={0 5 65 77},clip,width=\linewidth]{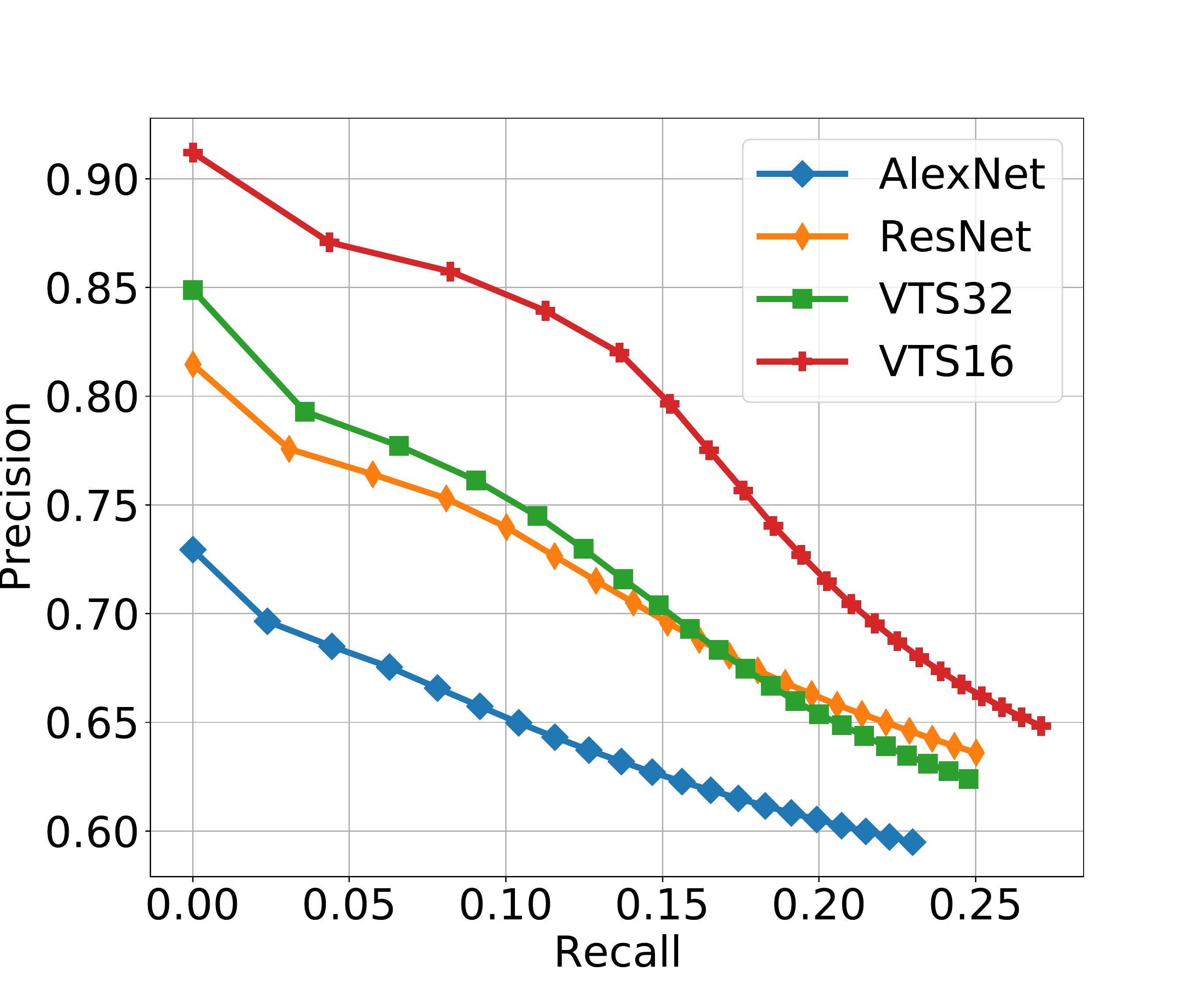} &
\includegraphics[trim={0 5 65 77},clip,width=\linewidth]{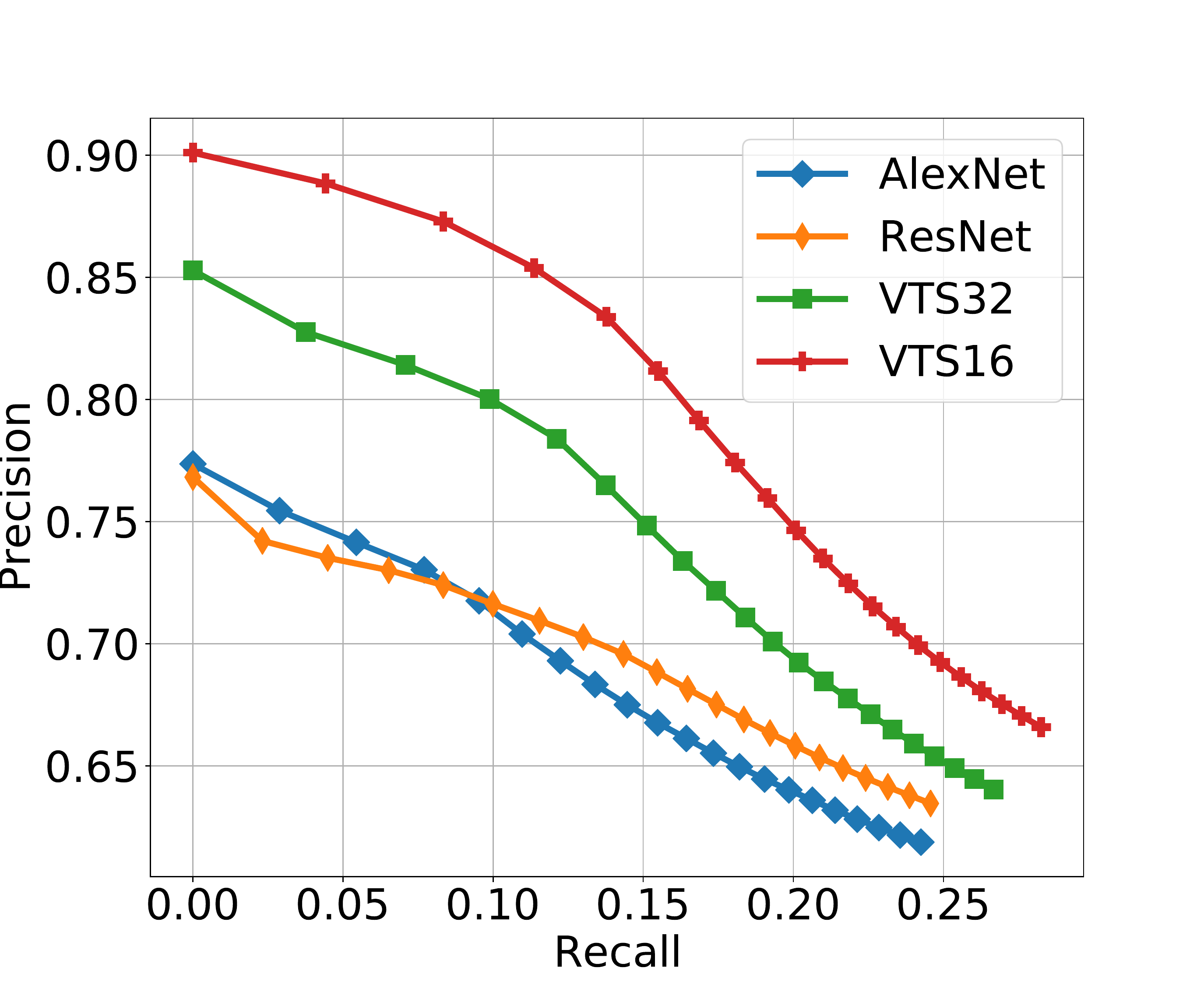}  &
\includegraphics[trim={0 5 65 77},clip,width=\linewidth]{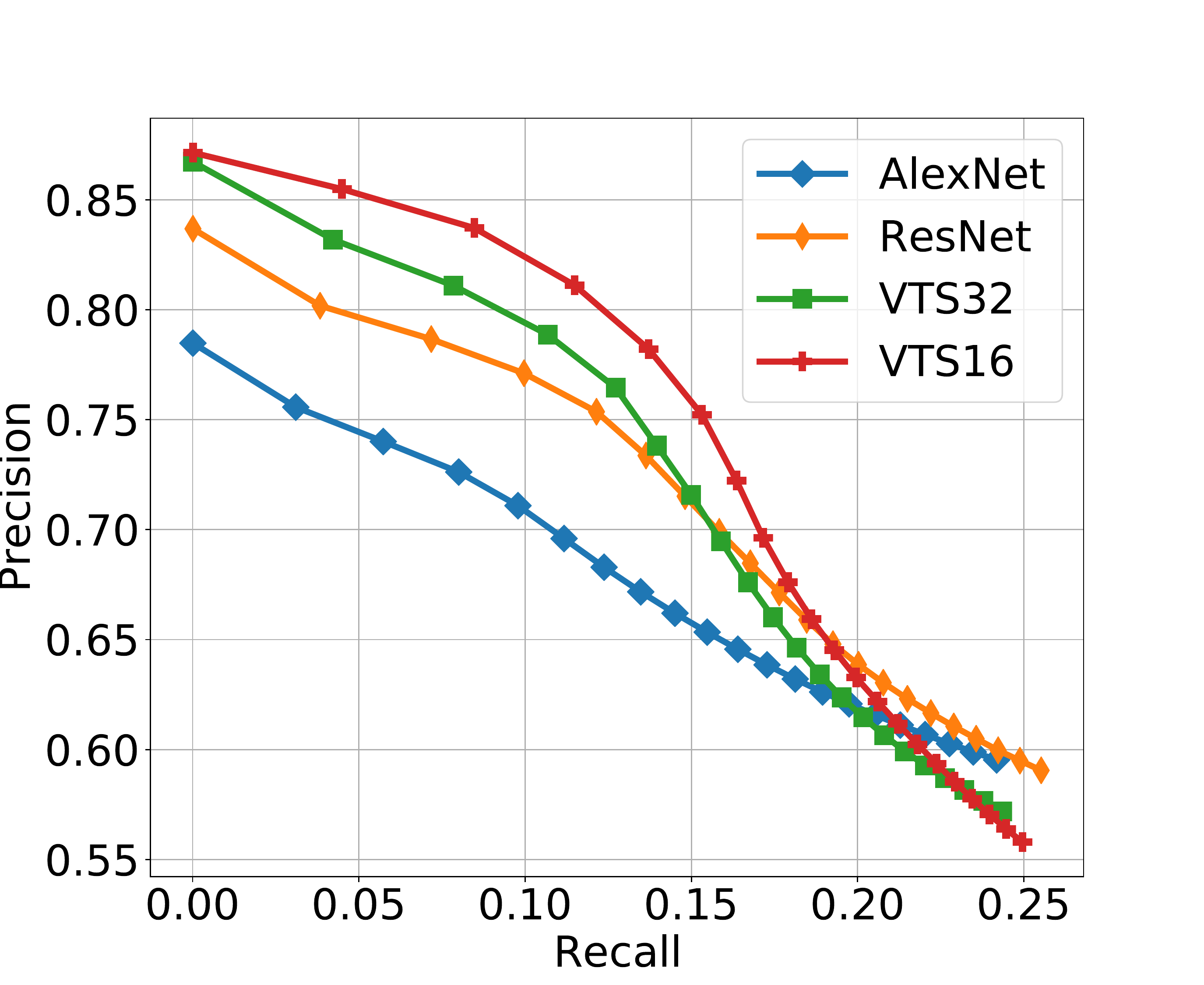}  &
\includegraphics[trim={15 5 65 77},clip,width=\linewidth]{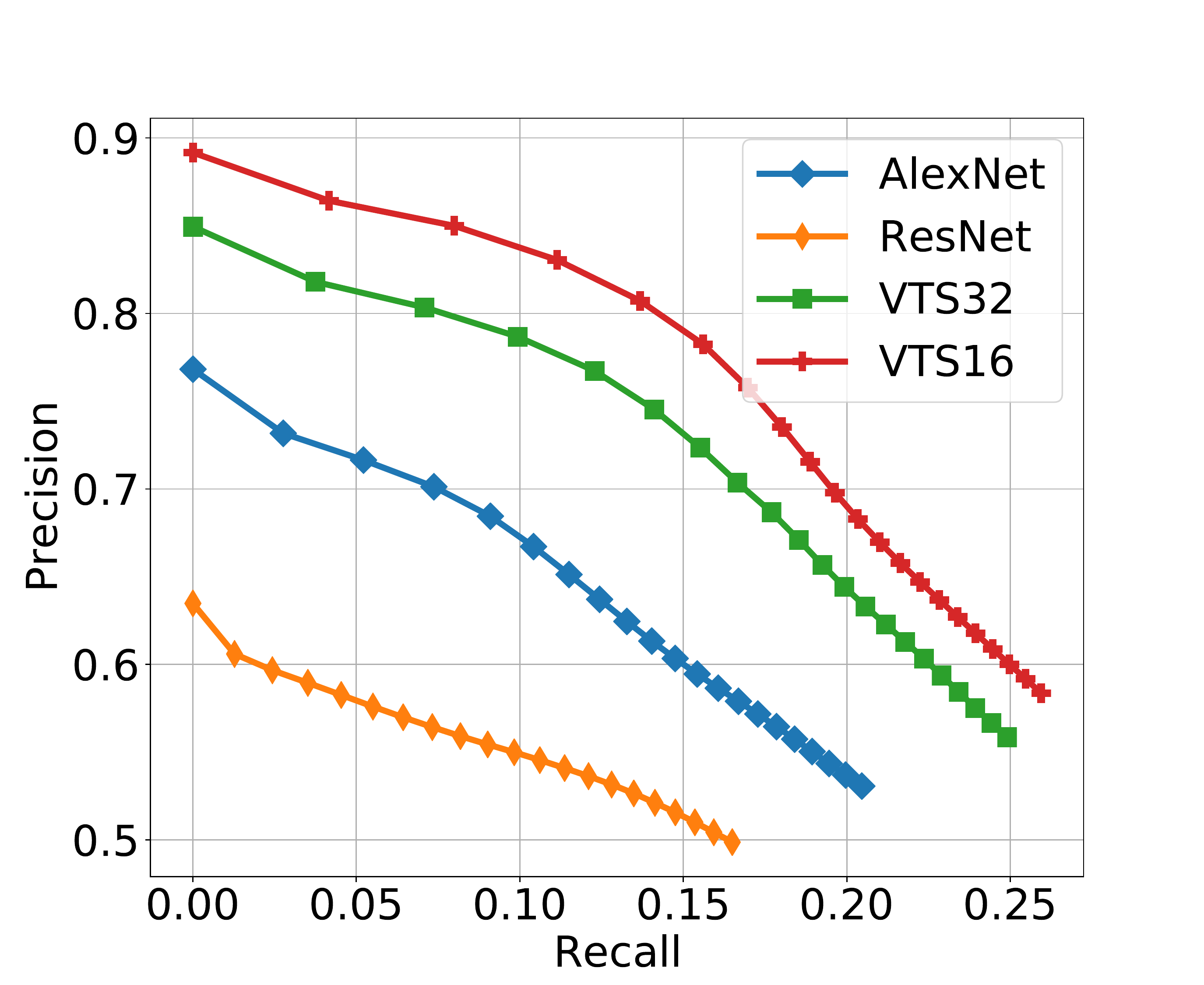}  &
\includegraphics[trim={0 5 65 77},clip,width=\linewidth]{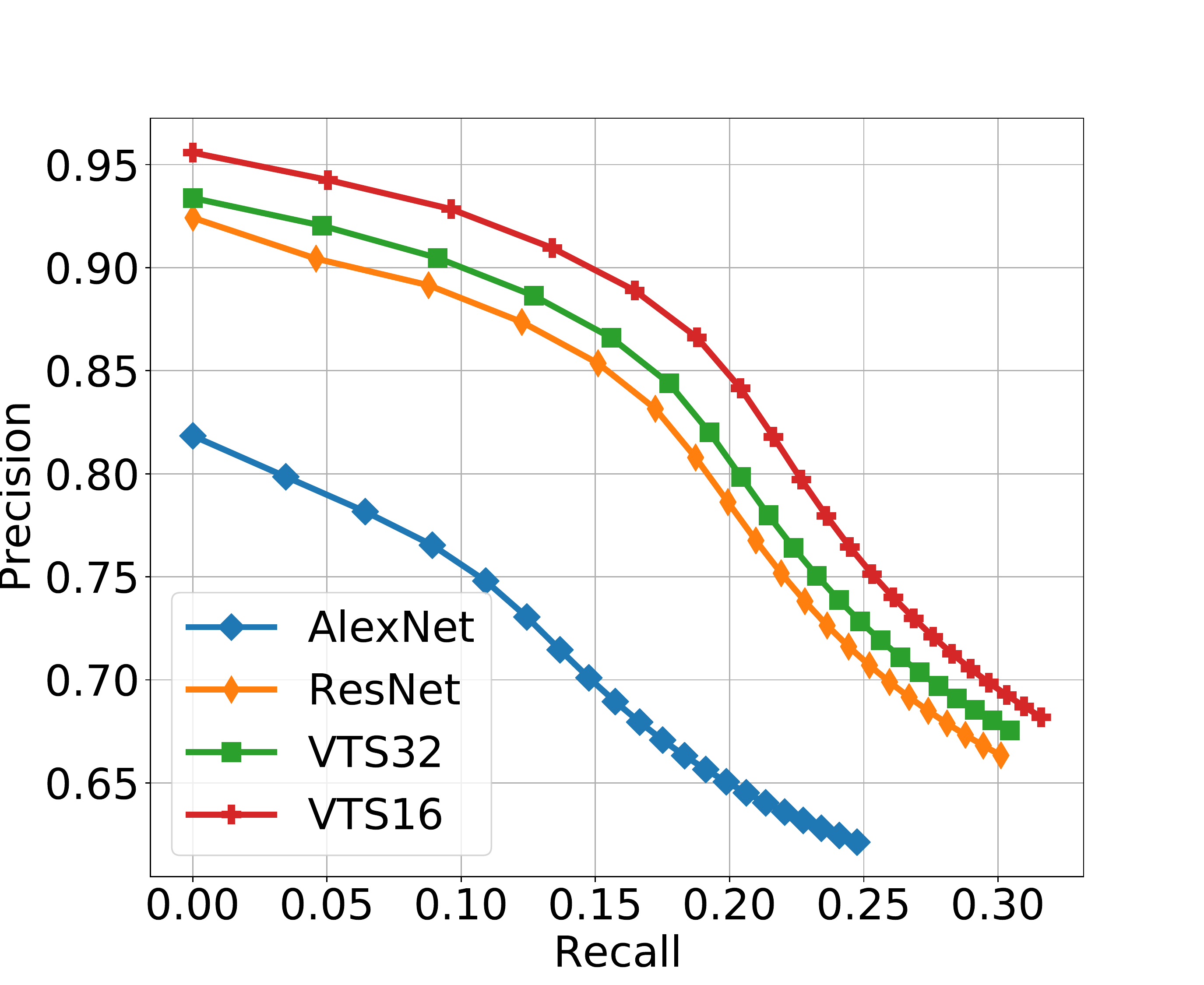}  &
\includegraphics[trim={0 5 65 77},clip,width=\linewidth]{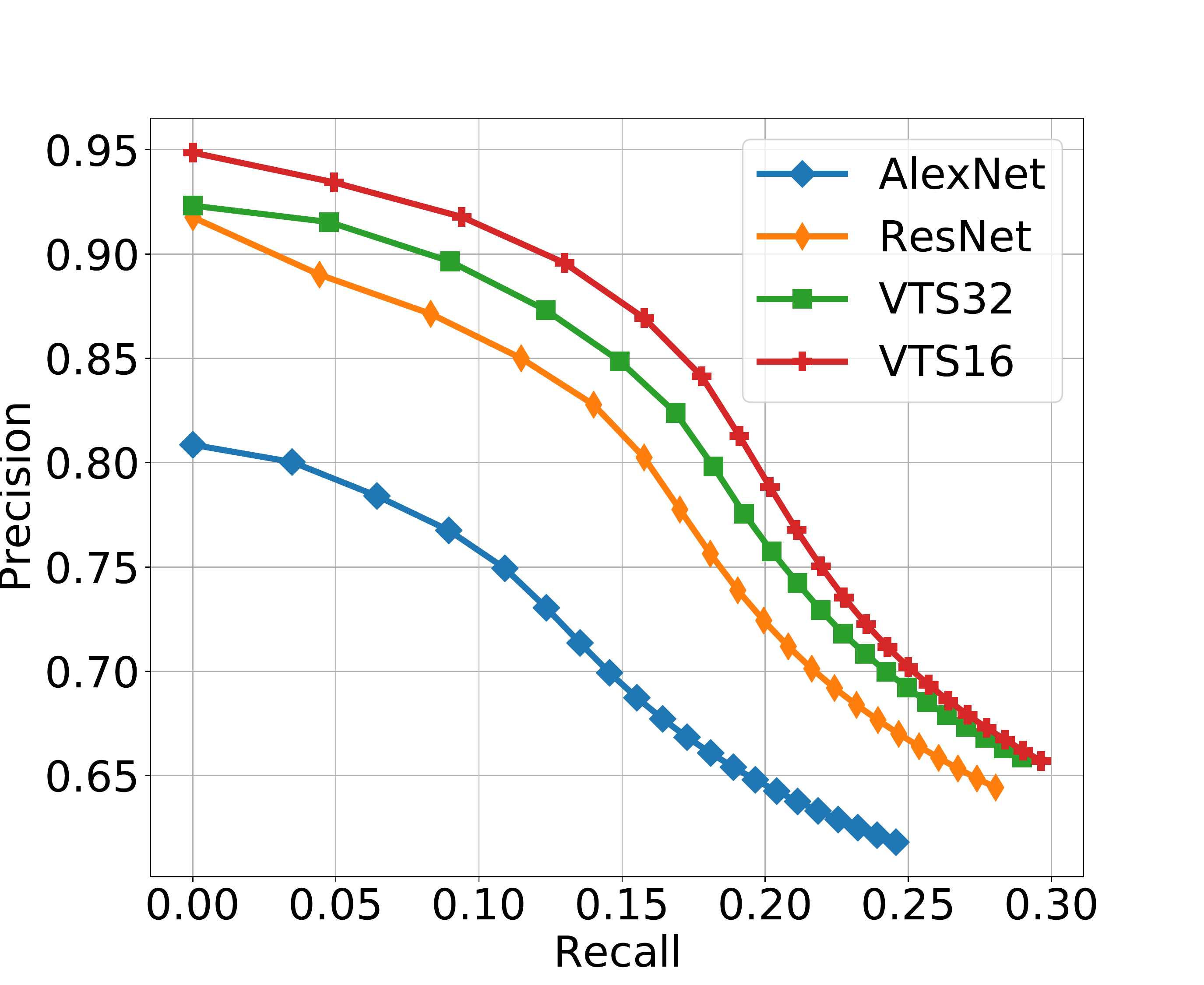} \\
\end{tabular}}
\caption{The RoC curves by using 64 bit hash code generated from AlexNet, ResNet, VTS32 and VTS16 backbone networks under DSH ($1^{st}$ column), HashNet ($2^{nd}$ column), GreedyHash ($3^{rd}$ column), IDHN ($4^{th}$ column), CSQ ($5^{th}$ column), and DPN ($6^{th}$ column) frameworks on CIFAR-10@54000 ($1^{st}$ row), CIFAR-10@All ($2^{nd}$ row), ImageNet@1000 ($3^{rd}$ row), NUS-Wide@5000 ($4^{th}$ row), and MS-COCO@5000 ($5^{th}$ row) datasets.}
\label{fig:roc}
\end{figure*}

\section{Experimental Results and Analysis}
This section presents the detailed experimental results over four benchmark datasets using the proposed VTS models under six retrieval frameworks. The ROC plots are also presented. The results are also compared with the state-of-the-art methods. An ablation study is also performed on the utilization of the features of the transformer encoder.
We perform the experiments using PyTorch deep learning library on a system with 24 GB NVIDIA RTX GPU.

\begin{table*}[!t]
\caption{The retrieval results comparison in terms of mAP (\%) with state-of-the-art results on CIFAR-10, ImageNet, NUS-Wide and MS-COCO datasets for 16, 32 and 64 bit hash codes. The results of the proposed VTS16 model under CSQ framework are used for comparison. The $\star$, $\dagger$ and $\ddagger$ symbols indicate that the corresponding results are taken from DCH \cite{dch}, DFH \cite{dfh} and DPN \cite{dpn}, respectively. The $+$ symbol indicates that all 81 classes of NUS-Wide dataset are used for these results. The best results are highlighted in bold.}
    \centering
    \resizebox{\textwidth}{!}{
    \begin{tabular}{m{0.283\textwidth}|m{0.026\textwidth}m{0.026\textwidth}m{0.033\textwidth}|m{0.026\textwidth}m{0.026\textwidth}m{0.036\textwidth}|m{0.026\textwidth}m{0.026\textwidth}m{0.033\textwidth}|m{0.026\textwidth}m{0.026\textwidth}m{0.033\textwidth}|m{0.026\textwidth}m{0.026\textwidth}m{0.033\textwidth}m{0.026\textwidth}m{0.026\textwidth}m{0.033\textwidth}}
    \hline
	Method & \multicolumn{3}{|c|}{CIFAR10@54000} & \multicolumn{3}{c|}{CIFAR10@All} & \multicolumn{3}{c|}{ImageNet@1000} & \multicolumn{3}{c|}{NUS-Wide@5000} & \multicolumn{3}{c}{MS-COCO@5000} \\ \cline{2-16}
	& 16b & 32b & 64b & 16b & 32b & 64b & 16b & 32b & 64b & 16b & 32b & 64b & 16b & 32b & 64b \\ \hline
	
	DSH (CVPR) \cite{dsh} & - & - & - & - & 66.1 & & - & - & - & - & 55.8 & - & - & - & - \\
    DHN (AAAI) \cite{dhn} & 69.3$^\star$ & 64.5$^\star$ & 58.8$^\star$ & - & - & - & 31.1$^\dagger$ & 47.2$^\dagger$ & 57.3$^\dagger$ & - & 74.8 & - & 67.5$^\star$ & 66.8$^\star$ & 41.9$^\star$ \\
    HashNet (ICCV) \cite{hashnet} & 74.8$^\star$ & 77.8$^\star$ & 62.6$^\star$ & 64.3$^\ddagger$ & 67.5$^\ddagger$ & 68.7$^\ddagger$ & 50.6 & 63.1 & 68.4 & 66.2$^+$ & 69.9$^+$ & 71.6$^+$ & 68.7 & 71.8 & 73.6 \\
    GreedyHash (NIPS) \cite{greedyhash} & - & - & - & - & 81.0 & - & 62.5 & 66.2 & 68.8 & - & - & - & - & - & -	\\
    DCH (CVPR) \cite{dch} & 79.0 & 79.8 & 79.4 & - & - & - & - & - & - & 74.0$^+$ & 77.2$^+$ & 71.2$^+$ & 70.1 & 75.8 & 70.1 \\
    DFH (BMVC) \cite{dfh} & - & - & - & - & 83.1 & - & 59.0 & 69.7 & 74.7 & - & 83.3 & - & - & - & - \\
    CSQ (CVPR) \cite{csq} & - & - & - & - & - & - & 85.1 & 86.5 & 87.3 & 81.0 & 82.5 & 83.9 & \textbf{79.6} & 83.8 & 86.1 \\
    DPN (IJCAI) \cite{dpn} & - & - & - & 77.4 & 80.3 & 81.2 & 60.6 & 69.3 & 72.9 & 81.0 & 83.2 & 83.9 & - & - & - \\
    TransHash \cite{transhash} &	90.8 & 91.1 & 91.7 & - & - & - & 78.5 & 87.3 & 89.2 & 72.6$^+$ & 73.9$^+$ & 74.9$^+$ & - & - & - \\\hline
    \textbf{VTS16-CSQ (Proposed)} & \textbf{97.9} & \textbf{97.9} & \textbf{97.6} & \textbf{97.6} & \textbf{97.7} & \textbf{97.6} & \textbf{90.6} & \textbf{91.5} & \textbf{91.8} & \textbf{81.9} & \textbf{84.6} & \textbf{85.3} & 77.9 & \textbf{87.4} & \textbf{91.1} \\\hline
    \end{tabular}}
    \label{table:results_comparison}
\end{table*}

\begin{table*}[!t]
\caption{The retrieval results in terms of mAP in \% for 16, 32 and 64 bit hash codes using only cls features and all features of the proposed VTS32 and VTS16 models under DSH, HashNet, GreedyHash, IDHN, CSQ and DPN frameworks on CIFAR-10, ImageNet, NUS-Wide and MS-COCO datasets. The best results are highlighted in bold.}
    \centering
    \resizebox{\textwidth}{!}{
    \begin{tabular}{m{0.098\textwidth}|m{0.026\textwidth}m{0.026\textwidth}m{0.026\textwidth}|m{0.026\textwidth}m{0.026\textwidth}m{0.026\textwidth}|m{0.026\textwidth}m{0.026\textwidth}m{0.026\textwidth}|m{0.026\textwidth}m{0.026\textwidth}m{0.026\textwidth}|m{0.026\textwidth}m{0.026\textwidth}m{0.026\textwidth}|m{0.026\textwidth}m{0.026\textwidth}m{0.026\textwidth}}
    \hline
	Backbone & \multicolumn{3}{|c|}{DSH} & \multicolumn{3}{c|}{HashNet} & \multicolumn{3}{c|}{GreedyHash} & \multicolumn{3}{c|}{IDHN} & \multicolumn{3}{c|}{CSQ} & \multicolumn{3}{c}{DPN} \\ \cline{2-19}
	Network & 16b & 32b & 64b & 16b & 32b & 64b & 16b & 32b & 64b & 16b & 32b & 64b & 16b & 32b & 64b & 16b & 32b & 64b \\ \hline
	
	\multicolumn{19}{c}{Results on CIFAR-10 dataset (mAP@54000 in \%)}\\\hline
	VTS32 (cls) & \textbf{93.2} & \textbf{94.6} & 95.1 & 95.6 & 96.4 & 95.5 & 94.2 & 93.7 & \textbf{94.8} & 94.8 & 94.8 & 94.6 & 95.2 & 95.1 & \textbf{95.6} & \textbf{95.8} & 95.0 & 93.8 \\
	VTS32 (all) & 93.0 & \textbf{94.6} & \textbf{95.3} & \textbf{96.2} & \textbf{97.1} & \textbf{95.6} & \textbf{94.8} & \textbf{95.6} & 94.3 & \textbf{95.9} & \textbf{95.8} & \textbf{96.0} & \textbf{96.0} & \textbf{96.1} & 95.2 & 95.1 & \textbf{95.6} & \textbf{95.0} \\\hline
	VTS16 (cls) & 96.2 & 96.9 & 97.0 & \textbf{97.8} & 97.3 & 97.9 & 94.8 & 95.9 & 97.2 & 97.1 & 97.1 & 97.1 & 97.8 & 97.6 & 97.1 & 96.8 & 96.2 & 96.4 \\
	VTS16 (all) & \textbf{98.0} & \textbf{97.4} & \textbf{97.4} & \textbf{97.8} & \textbf{98.2} & \textbf{98.3} & \textbf{97.0} & \textbf{97.0} & \textbf{98.0} & \textbf{97.6} & \textbf{97.5} & \textbf{98.4} & \textbf{97.9} & \textbf{97.9} & \textbf{97.6} & \textbf{97.4} & \textbf{96.7} & \textbf{97.5} \\\hline
	
	\multicolumn{19}{c}{Results on CIFAR-10 dataset (mAP@All in \%)}\\\hline
	VTS32 (cls) & 92.7 & \textbf{96.3} & \textbf{96.2} & 95.6 & 95.8 & 95.9 & 95.4 & \textbf{96.0} & \textbf{96.3} & 95.0 & \textbf{95.7} & \textbf{95.4} & 96.1 & 95.4 & \textbf{96.7} & 94.2 & 94.1 & 92.9 \\
	VTS32 (all) & \textbf{93.7} & 95.0 & 95.4 & \textbf{96.2} & \textbf{95.9} & \textbf{96.6} & \textbf{95.6} & 94.9 & 95.8 & \textbf{95.5} & 95.5 & 94.3 & \textbf{96.2} & \textbf{97.0} & 96.1 & \textbf{96.3} & \textbf{95.5} & \textbf{95.2} \\\hline
	VTS16 (cls) & 97.0 & 97.8 & 97.5 & 97.5 & 97.7 & \textbf{98.3} & 95.5 & 96.7 & 96.4 & 97.7 & 97.3 & 97.3 & \textbf{97.9} & 97.1 & \textbf{97.6} & 96.1 & 96.1 & 95.6 \\
	VTS16 (all) & \textbf{97.4} & \textbf{98.1} & \textbf{98.3} & \textbf{98.1} & \textbf{97.9} & 98.2 & \textbf{97.0} & \textbf{97.7} & \textbf{97.5} & \textbf{97.8} & \textbf{98.0} & \textbf{97.9} & 97.6 & \textbf{97.7} & \textbf{97.6} & \textbf{96.8} & \textbf{96.5} & \textbf{97.7} \\\hline
	
	\multicolumn{19}{c}{Results on ImageNet dataset (mAP@1000 in \%)}\\\hline
	VTS32 (cls) & \textbf{77.5} & \textbf{85.5} & \textbf{88.0} & \textbf{80.7} & \textbf{83.1} & \textbf{85.3} & 84.9 & \textbf{87.0} & 86.6 & \textbf{80.6} & \textbf{79.6} & \textbf{71.3} & 84.2 & 86.4 & 86.7 & \textbf{84.9} & 87.1 & 86.7 \\
	VTS32 (all) & 51.8 & 80.6 & 86.0 & 61.1 & 81.2 & 84.4 & \textbf{86.5} & 86.5 & \textbf{88.0} & 71.0 & 72.7 & 66.8 & \textbf{86.6} & \textbf{88.3} & \textbf{88.6} & 84.5 & \textbf{88.1} & \textbf{88.7} \\\hline
	VTS16 (cls) & 81.0 & 90.1 & 91.1 & \textbf{87.2} & \textbf{88.4} & 89.3 & 89.4 & 90.3 & 90.2 & \textbf{88.6} & \textbf{84.4} & \textbf{79.2} & 89.9 & 89.1 & 89.0 & 88.6 & 90.6 & 90.4 \\
	VTS16 (all) & \textbf{85.6} & \textbf{90.8} & \textbf{91.7} & 81.7 & 88.3 & \textbf{89.5} & \textbf{90.3} & \textbf{91.1} & \textbf{91.6} & 80.5 & 67.2 & 64.2 & \textbf{90.6} & \textbf{91.5} & \textbf{91.8} & \textbf{90.2} & \textbf{91.4} & \textbf{91.8}\\\hline
	
	\multicolumn{19}{c}{Results on NUS-Wide dataset (mAP@5000 in \%)}\\\hline
	VTS32 (cls) & 77.1 & \textbf{81.2} & 81.4 & \textbf{78.5} & 83.6 & \textbf{86.3} & \textbf{76.5} & \textbf{78.4} & \textbf{80.6} & \textbf{83.7} & \textbf{84.4} & 83.9 & 80.9 & 84.9 & \textbf{86.2} & \textbf{80.4} & 83.2 & \textbf{86.3} \\
	VTS32 (all) & \textbf{79.6} & 80.5 & \textbf{82.9} & 77.6 & \textbf{83.8} & 86.2 & 75.8 & 78.2 & 79.7 & 83.3 & \textbf{84.1} & 84.1 & \textbf{82.4} & \textbf{85.5} & 86.1 & 80.2 & \textbf{84.1} & 86.1 \\\hline
	VTS16 (cls) & \textbf{80.1} & 81.0 & 81.4 & 79.1 & 85.0 & 87.2 & \textbf{76.2} & \textbf{76.9} & 78.5 & 83.9 & 85.0 & 83.8 & 79.9 & 83.9 & \textbf{85.3} & 77.0 & 83.2 & \textbf{85.3} \\
	VTS16 (all) & 79.8 & \textbf{81.7} & \textbf{82.9} & \textbf{79.2} & \textbf{85.2} & \textbf{87.3} & \textbf{76.2} & \textbf{76.9} & \textbf{78.6} & \textbf{84.1} & \textbf{85.2} & \textbf{85.1} & \textbf{81.9} & \textbf{84.6} & \textbf{85.3} & \textbf{79.7} & \textbf{83.5} & 84.6 \\\hline
	
	\multicolumn{19}{c}{Results on MS-COCO dataset (mAP@5000 in \%)}\\\hline
	VTS32 (cls) & \textbf{73.0} & \textbf{75.8} & \textbf{77.9} & \textbf{70.7} & 75.8 & 79.9 & 73.7 & 77.2 & 79.2 & 75.3 & \textbf{81.1} & 78.3 & 70.2 & 83.0 & 85.9 & 71.3 & 82.7 & 85.8 \\
	VTS32 (all) & 71.9 & \textbf{75.8} & 76.9 & 70.6 & \textbf{76.8} & \textbf{80.6} & \textbf{75.5} & \textbf{79.6} & \textbf{80.3} & \textbf{76.3} & 79.9 & \textbf{78.4} & \textbf{77.0} & \textbf{86.5} & \textbf{88.9} & \textbf{75.3} & \textbf{84.5} & \textbf{88.0} \\\hline
	VTS16 (cls) & 78.6 & 81.8 & 84.3 & 75.1 & 82.7 & 84.2 & 77.0 & 79.9 & 80.7 & 77.4 & \textbf{83.8} & 69.3 & 72.5 & 83.0 & 86.5 & 73.5 & 78.6 & 86.6 \\
	VTS16 (all) & \textbf{80.5} & \textbf{84.3} & \textbf{85.0} & \textbf{75.2} & \textbf{82.8} & \textbf{86.5} & \textbf{78.0} & \textbf{81.2} & \textbf{82.3} & \textbf{77.8} & 82.8 & \textbf{83.0} & \textbf{77.9} & \textbf{87.4} & \textbf{91.1} & \textbf{78.5} & \textbf{85.8} & \textbf{89.7} \\\hline
    \end{tabular}}
    \label{table:ablation_results}
\end{table*}

\subsection{Experimental Results}
The experimental results in terms of mAP (\%) are summarized in Table \ref{table:retrieval_results} for image retrieval on CIFAR10, ImageNet, NUS-Wide and MS-COCO datasets. The results of the proposed VTS32 and VTS16 backbone networks are compared with the AlexNet and ResNet50 backbone networks. The results are computed using 16 bit (16b), 32 bit (32b) and 64 bit (64b) hash codes under six retrieval frameworks, namely DSH, HashNet, GreedyHash, IDHN, CSQ and DPN. It can be noted that the proposed VTS16 outperforms the other backbone networks in all the settings on CIFAR10 and MS-COCO datasets and also better in most of the cases on ImageNet and NUS-Wide datasets. Moreover, the performance of VTS32 is also superior to AlexNet and ResNet50 in the majority of cases. Following are the observations:
\begin{itemize}
    \item On an average the mAP of VTS16 backbone network is improved by $26.65 \%$, $54.68 \%$, $3.15 \%$, and $20.56 \%$ as compared to AlexNet backbone network and by $23.74 \%$, $31.50 \%$, $2.67 \%$, and $12.78 \%$ as compared to ResNet50 backbone network on CIFAR10, ImageNet, NUS-Wide and MS-COCO datasets, respectively.
    \item On an average the mAP of VTS32 backbone network is improved by $23.86 \%$, $42.94 \%$, $3.03 \%$, and $15.14 \%$ as compared to AlexNet backbone network and by $21.01 \%$, $21.51 \%$, $2.56 \%$, and $7.71 \%$ as compared to ResNet50 backbone network on CIFAR10, ImageNet, NUS-Wide and MS-COCO datasets, respectively.
\end{itemize}

We also compare the precision recall ROC plots of AlexNet, ResNet50, VTS32 and VTS16 backbone networks for 64 bit hash code in Fig. \ref{fig:roc}. The ROC plots corresponding to DSH, HashNet, GreedyHash, IDHN, CSQ and DPN retrieval frameworks are drawn in $1^{st}$ to $6^{th}$ columns, respectively. The $1^{st}$ to $5^{th}$ rows correspond to ROC plots on CIFAR10@54000, CIFAR10@All, ImageNet@1000, NUS-Wide@5000 and MS-COCO@5000 datasets in order. It is evident from these plots that both the VTS16 and VTS32 backbone networks perform better than the AlexNet and ResNet50 backbone networks in all the cases except on NUS-Wide under the GreedyHash framework. It is also observed that VTS16 is superior in the majority of cases.

\subsection{Comparison with Existing Results}
Table \ref{table:results_comparison} presents the results of the proposed vision transformer hashing with 16 patch size under CSQ retrieval framework (i.e., VTS16-CSQ) and the results of the state-of-the-art retrieval models, including DSH \cite{dsh}, DHN \cite{dhn}, HashNet \cite{hashnet}, GreedyHash \cite{greedyhash}, DCH \cite{dch}, DFH \cite{dfh}, CSQ \cite{csq}, DPN \cite{dpn} and TransHash \cite{transhash}. Note that TransHash is a recent siamese transformer model for image retrieval. Following are the observations from these results comparison:
\begin{itemize}
    \item It is observed that the proposed hashing technique outperforms the existing hashing approaches on CIFAR10, ImageNet, NUS-Wide and MS-COCO datasets for 16 bit, 32 bit and 64 bit hash codes.
    \item It is also evident that the pre-training of vision transformer models plays an important role as the VTS16-CSQ model clearly swaps the TransHash model on CIFAR10, ImageNet and NUS-Wide datasets.
\end{itemize}

\subsection{Ablation Study}

We use \textit{all} features ($TE$) of output of the transformer encoder in the hash layer in the proposed VTS32 and VTS16 models in the earlier results. In order to analyze the impact of features, the results using only class (\textit{cls}) features of the proposed transformer model are compared with all the features in Table \ref{table:ablation_results} as an ablation study under different retrieval frameworks. Note that the \textit{cls} feature represents the output features of the vision transformer corresponding to the output unit which was used for MLP Head originally as shown in Fig. \ref{fig:vts}.  The VTS32 (cls) and VTS32 (all) refer to the computed hash code from only \textit{cls} features and \textit{all} features, respectively, while the patch-size is considered as 32. Similarly, the VTS16 (cls) and VTS16 (all) refer to the computed hash code from only \textit{cls} features and \textit{all} features, respectively, while the patch-size is considered as 16. It can be observed that the performance of VTS16 is superior in most of the cases with \textit{all} features. However, the performance of VTS32 using \textit{cls} features is generally better on ImageNet dataset as the pretrained vision transformer model was originally trained on the same dataset. Overall, the utilization of \textit{all} features to compute the VTS hash code is advised in order to improve the generalization capability of the network.

\section{Conclusion}
In this paper, we propose a vision transformer based hashing (VTS) framework for image retrieval using the pretrained ViT model as the backbone network. 
We train the VTS models under six retrieval frameworks
to extract the hash code.
The training is performed in an end-to-end fashion using the objective function of the corresponding retrieval framework.
The results are compared on CIFAR10, ImageNet, NUS-Wide and MS-COCO datasets. It is noticed that the proposed VTS backbone outperforms the AlexNet and ResNet backbones under different retrieval frameworks. Moreover, the performance of the proposed model is superior to the state-of-the-art hashing techniques, including transformer based technique. We also observe that the CSQ framework is better suited with the proposed VTS model for image retrieval. It is also found that the consideration of all the features of the vision transformer's output increases the generalization capability of the model. Thus, it can be concluded that the vision transformer based hashing significantly boosts the performance of image retrieval due to it's learning capability of discriminative features.

\section*{Acknowledgement}
This research was funded by the Global Innovation and Technology Alliance (GITA) on Behalf of Department of Science and Technology (DST), Govt. of India under India-Taiwan joint project with Project Code GITA/DST/TWN/P-83/2019.

{\small
\bibliographystyle{ieee_fullname}
\bibliography{References}
}

\end{document}